\documentclass[runningheads]{llncs}
\usepackage{cite}  
\usepackage{xcolor}
\usepackage{siunitx}
\usepackage{booktabs}
\usepackage{graphicx}        
\usepackage{textcomp} 
\usepackage{multirow}
\usepackage{hyperref}
\usepackage{makecell}  
\usepackage{tabularx}
\usepackage{subcaption}
\usepackage[T1]{fontenc}
\usepackage{amsmath, amssymb, amsfonts}
\usepackage[ruled,linesnumbered]{algorithm2e} 

\begin{document}

\title{CoAction: Cross-task Correlation-aware Pareto Set Learning}

\author{Xinyue Chen\inst{1}\and
Yingxuan Liang\inst{1}\and
Yiqin Huang\inst{1}\and
Chikai Shang\inst{2}\and
Hai-Lin Liu\inst{1}\and
Fangqing Gu\inst{1}$^*$}

\authorrunning{X. Chen et al.}

\institute{Guangdong University of Technology, Guangzhou, China \\
\and Xiamen University, Xiamen, China \\
\email{fqgu@gdut.edu.cn}}

\maketitle

\begin{abstract}
Pareto set learning (PSL) is an emerging paradigm in multi-objective optimization that trains neural networks to map preference vectors to Pareto optimal solutions. However, existing PSL methods primarily focus on solving a single multi-objective optimization problem at a time. This limitation not only increases computational costs in multi-objective multitask optimization scenarios by requiring a separate model for each task, but also fails to exploit the inter-task correlations across tasks. To address this, we propose a \underline{C}r\underline{o}ss-t\underline{A}sk \underline{c}orrela\underline{tion}-aware Pareto Set Learning (CoAction) framework, which leverages task-aware transformer to handle multiple tasks simultaneously. Specifically, by assigning task-specific embedding vectors to individual tasks, the model effectively distinguishes between tasks while facilitating knowledge sharing among them. We utilize a Transformer encoder as the backbone architecture to leverage its self-attention mechanism for capturing complex task dependencies. The proposed approach is evaluated on comprehensive multitask test suites covering both benchmark problems and real-world applications, demonstrating effectiveness and competitive performance in Hypervolume, Range, and Sparsity.

\keywords{Pareto set learning \and task embedding \and knowledge sharing \and self-attention mechanism}
\end{abstract}

\section{Introduction}
Multi-objective optimization (MOO) aims to simultaneously optimize multiple conflicting objectives~\cite{MultiobjectiveOU, Nonlinear}, yielding a set of trade-off solutions known as the Pareto set (PS), whose image in objective space constitutes the Pareto front (PF)~\cite{Comparison}. Many practical applications require Pareto optimal solutions to be generated in real time to support rapid switching between trade-off preferences. However, traditional evolutionary methods such as NSGA-II~\cite{AFA}, MO-CMA-ES~\cite{CovarianceMA}, SMS-EMOA~\cite{SMS}, and MOEA/D~\cite{MOEADAM} produce only a finite number of optimal solutions per run and cannot respond to arbitrary preference without restarting the search, hindering real-time decision-making. To address this limitation, Lin et al.~\cite{ControllablePM} proposed Pareto set learning (PSL), which trains a neural network to approximate the entire PS conditioned on preference vectors, enabling real-time generation of Pareto optimal solutions for any given preference.

Despite these advances, a fundamental limitation of existing PSL methods is that they are designed to solve one optimization problem at a time, requiring an independent model to be trained for each problem. This paradigm incurs a computational cost that scales linearly with the number of problems and prevents the model from capturing and exploiting shared structure across related tasks. Such limitations become especially critical in practical settings where a large number of related optimization problems arise simultaneously and must be addressed efficiently. A natural and practically important example of such settings is multitask learning (MTL)~\cite{Rich1997Multitask, Yu2022MultiTaskLearning}, where multiple tasks are optimized jointly and their inherent shared structure offers an opportunity for cross-task knowledge sharing. Sener and Koltun~\cite{Sener2018MultiTaskLA} formally cast the joint optimization of multiple tasks as a MOO problem, providing a principled theoretical foundation that motivates the application and extension of PSL methods to MTL scenarios.

Extending PSL to handle multiple optimization problems simultaneously exposes several critical challenges. First, existing PSL methods treat each optimization problem independently, requiring a separate model per task and leading to significant computational inefficiency that scales linearly with the number of tasks~\cite{Zhang2023HypervolumeMA, Lin2022ParetoSL, Lin2022PSLN}. Moreover, this independent treatment fundamentally fails to exploit inter-task relationships, leaving the latent structural similarities across different optimization problems untapped and losing the regularization and generalization that shared representations could provide. These limitations motivate a unified framework that shares parameters across tasks while explicitly modeling inter-task relationships, enabling both scalability and improved generalization.

To address these challenges, we propose a Cross-task Correlation-aware Pareto Set Learning (CoAction) framework that learns Pareto optimal solution sets across multiple optimization problems within a single model. CoAction assigns unique task-specific embedding vectors to each optimization problem and employs explicit parameter sharing, enabling the model to distinguish between tasks and achieve cross-task knowledge sharing, thereby reducing computational costs. To explicitly model inter-task relationships, we adopt a Transformer encoder~\cite{Devlin2019BERTPO} as the backbone, whose self-attention mechanism captures complex global dependencies among tasks~\cite{Vaswani2017Attention} and supports scalable parallel PSL.

The main contributions of this paper are as follows.
\vspace{-5pt}
\begin{itemize}
	\item We propose a CoAction framework with a Transformer encoder backbone that leverages task embedding, shared network parameters, and self-attention mechanism to model inter-task relationships. This enables a single model to learn Pareto optimal solution sets for multiple optimization problems simultaneously, reducing computational costs while enhancing model expressiveness.
	\item We conduct extensive experiments on multitask test suites comprising benchmark problems and real-world engineering applications, demonstrating that the CoAction framework achieves competitive performance on hypervolume, range, and sparsity metrics.
\end{itemize}
\vspace{-5pt}

The remainder of this paper is organized as follows. Section~\ref{sec:Preliminaries} introduces the preliminaries of MOO, MTL, and the Transformer architecture. Section~\ref{sec:Multi-objective Multitask Optimization via Task Embedding} presents the proposed CoAction framework, including the task embedding scheme and the multitask collaborative training algorithm. Section~\ref{sec:Experiments} reports comprehensive experimental results on benchmark and real-world problems, along with ablation studies, extended evaluations, and statistical analyses. Finally, Section~\ref{sec:Conclusion} concludes the paper.

\section{Preliminaries} \label{sec:Preliminaries}
This section introduces the necessary background for the proposed method, covering MOO, MTL, and the Transformer architecture. 

\subsection{Multi-objective Optimization}
A MOO problem seeks to simultaneously minimize $m \geq 2$ objective functions over a decision space $\mathcal{X} \subseteq \mathbb{R}^n$:
\begin{equation}
    \min_{\mathbf{x} \in \mathcal{X}} F(\mathbf{x}) 
    = (f_1(\mathbf{x}), f_2(\mathbf{x}), \ldots, f_m(\mathbf{x})),
\end{equation}
where $\mathcal{Y} = F(\mathcal{X}) \subseteq \mathbb{R}^m$ denotes the image of $\mathcal{X}$ in the objective space.
\begin{definition} \textbf{Pareto Dominance.}
A solution $\mathbf{x}^{(a)}$ dominates $\mathbf{x}^{(b)}$, denoted 
$\mathbf{x}^{(a)} \prec \mathbf{x}^{(b)}$, if $f_i(\mathbf{x}^{(a)}) \leq f_i(\mathbf{x}^{(b)})$ for all $i \in \{1, 2, \ldots, m\}$ and $f_j(\mathbf{x}^{(a)}) < f_j(\mathbf{x}^{(b)})$ for at least one $j \in \{1, 2, \ldots, m\}$.
\end{definition}
\begin{definition} \textbf{Pareto Set and Pareto Front.}
A solution $\mathbf{x}^* \in \mathcal{X}$ is called a Pareto optimal solution if there exists no other solution that dominates it. The set of all Pareto optimal solutions is called the \textbf{PS}, and its image in objective space, $\mathcal{F} := F(PS)$, is called the \textbf{PF}.
\end{definition}
\begin{definition} \textbf{Hypervolume Indicator.}
Given a reference vector $\mathbf{r} \succeq \mathbf{y}^{\mathrm{nadir}}$, with $\mathbf{y}^{\mathrm{nadir}} = \max_{\mathbf{y} \in \mathrm{PF}} \mathbf{y}$ and $\mathbf{y}^{\mathrm{ideal}} = \min_{\mathbf{y} \in \mathrm{PF}} \mathbf{y}$ defined component-wise. The Hypervolume (HV) indicator of a solution set $A$ is defined as:
\begin{equation}
\mathcal{H}_{\mathbf{r}}(A) := \Lambda\!\left( \bigcup_{\mathbf{p} \in A} \left\{ \mathbf{q} \;\middle|\; \mathbf{p} \preceq \mathbf{q} \preceq \mathbf{r} \right\} \right),
\end{equation}
where $\Lambda(\cdot)$ denotes the Lebesgue measure. The HV quantifies the volume of the objective space dominated by $A$ and bounded by $\mathbf{r}$, and serves as a standard quality indicator for evaluating the approximated PF.
\end{definition}

\subsection{Multitask Learning}
MTL improves generalization by jointly training a single model on multiple related tasks, allowing shared statistical strength to benefit each individual task~\cite{Rich1997Multitask, Yu2022MultiTaskLearning}. Architecturally, MTL methods fall into two broad paradigms. Hard parameter sharing employs a common backbone with task-specific heads, acting as an implicit regularizer~\cite{Vandenhende2021MultiTaskLearning, Baxter2000AMO}, while soft parameter sharing maintains per-task parameters with explicit cross-task interaction~\cite{Ishan2016Cross-Stitch}. Beyond architectural choices, MTL has been formally cast as a MOO problem~\cite{Sener2018MultiTaskLA}. This formulation has inspired PSL-based approaches that condition a single model on preference vectors, enabling scalable inference of diverse trade-off solutions for MTL~\cite{Navon2021LearningTP, Liu2021ConflictAverseGD}.

\subsection{Transformer Architecture}
The Transformer~\cite{Vaswani2017Attention} relies on the self-attention mechanism to capture global dependencies across the input sequence. Given input embeddings $\mathbf{X} \in \mathbb{R}^{L \times d}$, the attention output is computed as follows:
\begin{equation}
    \mathrm{Attention}(\mathbf{Q}, \mathbf{K}, \mathbf{V})
    = \mathrm{softmax}\!\left(\frac{\mathbf{Q}\mathbf{K}^{\top}}{\sqrt{d_k}}\right)\mathbf{V},
\end{equation}
where $\mathbf{Q}$, $\mathbf{K}$, and $\mathbf{V}$ denote the query, key, and value matrices, respectively, each obtained via a linear projection of $\mathbf{X}$, and $d_k$ is the key dimensionality.

In multitask settings, conditioning a shared Transformer on task identity presents a central challenge. Two representative strategies have emerged: task embeddings~\cite{Liu2019MultiTaskDN} associate each task with a learnable vector prepended to the input or injected via cross-attention, while task query vectors~\cite{Ye2023TaskPrompterSM} introduce learnable task-specific tokens that attend over shared encoder representations, routing task-relevant information without modifying backbone weights. In the MOO setting, the preference vector $\boldsymbol{\lambda}$ naturally serves as a continuous task descriptor~\cite{Navon2021LearningTP, ControllablePM}, allowing a single model to approximate the full PF by varying $\boldsymbol{\lambda}$ at inference time.

\begin{figure*}[t]
    \centering
    \includegraphics[width=1.0\linewidth]{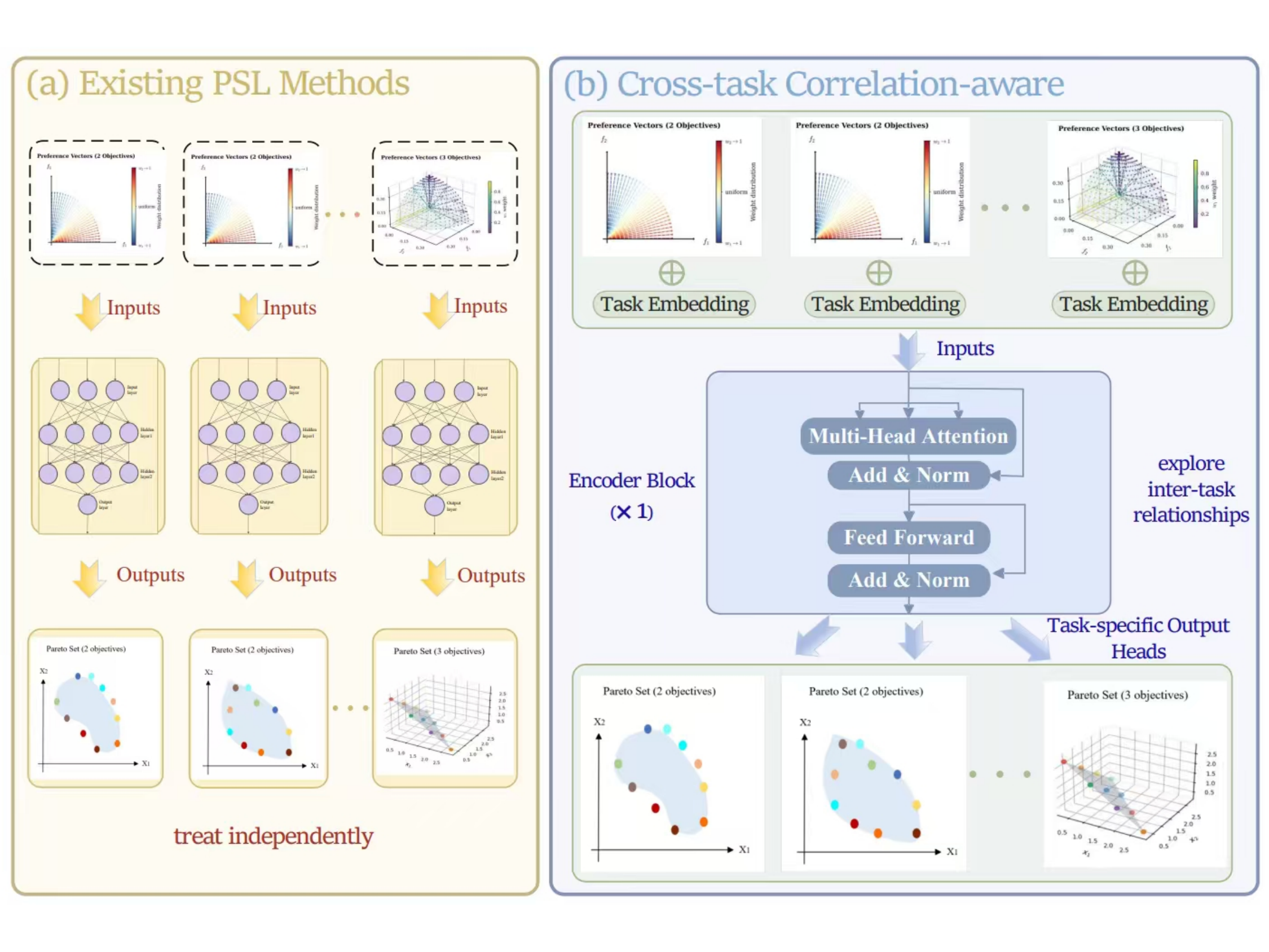}
    \caption{\textbf{Overall framework}. This figure illustrates the overall structure of our approach compared with the existing PSL methods. (a) Existing PSL methods solve each optimization problem independently, limiting efficiency and ignoring inter-task correlations. (b) The proposed CoAction framework integrates task embeddings with a Transformer encoder to simultaneously learn Pareto optimal solution sets for multiple problems, reducing computational cost while improving expressiveness.}
    \label{fig:method}
\end{figure*}

\section{Multi-objective Multitask Optimization via Task Embedding} \label{sec:Multi-objective Multitask Optimization via Task Embedding}
In this section, we present the proposed CoAction framework with task embedding. Specifically, we first establish the theoretical foundation and mathematical definition of task embedding, which serves as the core mechanism for task distinction. We then detail the implementation of the framework.

\subsection{Task Embedding}
In multitask scenarios, a central challenge is enabling the model to distinguish between tasks while capturing shared structure across them. We address this by deriving task representations analytically using sinusoidal functions. This eliminates the need for learnable embedding vectors~\cite{Liu2019MultiTaskDN}, which scale poorly with the number of tasks and fail to generalize when the number of tasks varies at inference time.

We use a task embedding scheme for multi-objective multitask optimization, where each task $t \in \{1, 2, \ldots, T\}$ is assigned a $d$-dimensional embedding vector $\mathbf{e}_t \in \mathbb{R}^{d}$ defined as:
\begin{equation}
    \begin{cases}
        \mathbf{e}_{(t, 2i)} = \sin\left(\dfrac{t}{50^{2i/d}}\right) \\
        \mathbf{e}_{(t, 2i+1)} = \cos\left(\dfrac{t}{50^{2i/d}}\right)
    \end{cases}.
\end{equation}
where $d$ is the embedding dimension, and $2i$, $2i+1$ index the even and odd dimensions, respectively. This construction is inspired by sinusoidal positional encoding~\cite{Vaswani2017Attention}. Whereas positional encoding $pos$ indexes a token's location within a sequence, the task embedding $t$ acts as a discrete identifier that distinguishes tasks. The resulting task embedding $\mathbf{e}_t$ is subsequently concatenated with the preference vector and a zero-padding vector to form the network input, as detailed in Section~\ref{sec:task_distinction}. 

\begin{figure*}[t]
    \centering
    \includegraphics[width=1.0\linewidth]{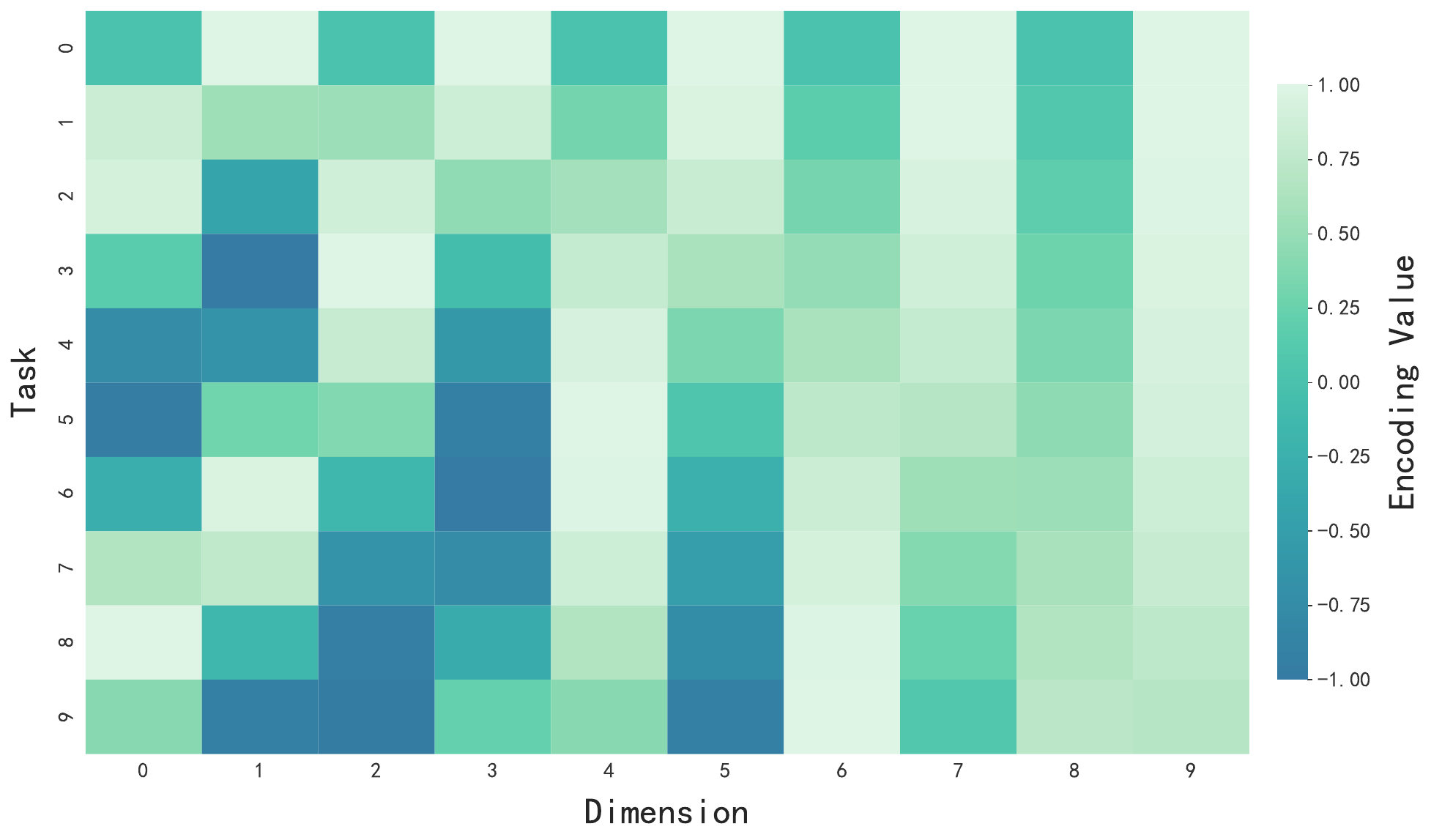}
    \caption{Task embedding vectors for $T = 10$ tasks with $d = 10$ for visualization purposes. Tasks with similar indices share greater embedding similarity,  while all tasks remain clearly distinguishable.}
    \label{fig:task_encoding}
\end{figure*}

Furthermore, these vectors can be organized into an embedding matrix $\mathbf{E} \in \mathbb{R}^{T \times d}$, whose $t$-th row is the embedding vector $\mathbf{e}_t$ of the task $t$. The matrix is given by 
\begin{equation}
    \mathbf{E} = ( \mathbf{e}_1; \mathbf{e}_2; \cdots; \mathbf{e}_T ).
\end{equation}

This sinusoidal construction provides two properties by design. First, smoothness, whereby tasks with similar indices receive similar embedding vectors to enable generalization across related tasks. Second, injectivity, whereby tasks with different indices remain sufficiently distinguishable to support task-specific learning. 

As illustrated in Fig.~\ref{fig:task_encoding}, each task receives a distinct embedding pattern while neighboring tasks share greater similarity, confirming both injectivity and smoothness in practice. The embedding dimension $d$ trades representational capacity against computational overhead. A larger $d$ enriches the task representations, but increases the number of model parameters. In this paper, we set $d=6$\footnote{Sensitivity analysis over $d \in \{2,4,6,8,10\}$ shows that HV and Range vary by less than 0.4\% across all values, while $d=6$ achieves notably lower Sparsity than $d \in \{2,4\}$ without the runtime overhead of $d \in \{8,10\}$.}.

\subsection{Proposed Algorithm}
In this section, we present the proposed CoAction framework. Existing PSL methods treat each optimization problem in isolation, lacking mechanisms for task identification, parameter sharing, and inter-task dependency modeling. To address these limitations, this framework introduces task embedding to distinguish tasks and shared neural network parameters to enable cross-task knowledge sharing. The model learns a continuous mapping from a low-dimensional preference space to a high-dimensional decision space. Given a preference vector, the network generates the corresponding Pareto optimal solution, enabling the model to capture both the optimization structure of individual problems and the shared knowledge across tasks.

\textbf{Preference Learning Mechanism}: PSL aims to learn a Pareto neural network model $\mathbf{x}_{\beta}(\cdot): \Theta \mapsto \mathbb{R}^n$, where $\beta$ denotes the model parameters, $n$ is the dimension of the decision variable space, and $\Theta = [0, \pi/2]^{m-1}$. This model maps a polar coordinate $\theta \in \Theta$ to a Pareto solution $\mathbf{x} \in \mathbb{R}^n$. Once trained, the Pareto neural network model $\mathbf{x}_{\beta}(\cdot)$ can generate approximate Pareto optimal solutions in real-time for any $\theta$.

Let \( \theta = (\theta_1, \ldots, \theta_{m-1}) \sim \text{Unif}(\Theta) \) be a polar coordinate vector sampled uniformly from \( \Theta \). The corresponding preference vector \( \boldsymbol{\lambda}(\mathbf{\theta}) \) represents the Cartesian coordinates of \( \mathbf{\theta} \) on the positive unit sphere \( \mathbb{S}^{m-1}_+ \). The mapping from \( \mathbf{\theta} \) to \( \boldsymbol{\lambda}(\mathbf{\theta}) \) is defined as:

\begin{equation}
	\begin{cases}
		\lambda_1(\theta) = \sin \theta_1 \sin \theta_2 \cdots \sin \theta_{m-1} \\
		\lambda_2(\theta) = \sin \theta_1 \sin \theta_2 \cdots \cos \theta_{m-1} \\
		\vdots \\
		\lambda_m(\theta) = \cos \theta_1
	\end{cases}.
\end{equation}

For bi-objective optimization problems ($m=2$), the preference vector is $\lambda(\theta) = (\sin \theta_1, \cos \theta_1)^{\mathrm{T}}$ with $\theta_1 \in [0, \pi/2]$. For tri-objective optimization problems ($m=3$), the preference vector is $\lambda(\theta) = (\sin \theta_1 \sin \theta_2, \sin \theta_1 \cos \theta_2, \cos \theta_1)^{\mathrm{T}}$ with $\theta_1, \theta_2 \in [0, \pi/2]$. This parameterization ensures $|\lambda(\theta)|_2 = 1$, and when $\theta$ is uniformly sampled, produces preference vectors that are evenly distributed over the positive orthant of the unit hypersphere, promoting solution diversity and training stability. During training, the network is optimized over randomly sampled preference vectors to learn a continuous mapping from the preference space to the PS.

\label{sec:task_distinction}
\textbf{Task Distinction Mechanism}: To enable the model to recognize and differentiate among multiple tasks, we introduce a task embedding mechanism that assigns a unique vector representation to each task. The task embedding is concatenated with the preference vector $\lambda(\theta)$ to form the input to the neural network $\mathbf{x}_{\beta}$, thereby allowing the model to generate task-specific decision variables. The uniqueness of each embedding preserves task independence, while the smooth and continuous variation across encodings facilitates cross-task knowledge sharing.
	
For each training sample, the network input $V$ consists of three components: task embedding $\mathbf{e}_t$, the preference vector $\boldsymbol{\lambda}(\theta)$, and a zero-padding vector that ensures a unified input dimensionality. Formally, the input is defined as:
\begin{equation}
		\mathbf{V} = [\mathbf{e}_t, \boldsymbol{\lambda}(\theta), \mathbf{0}_{d_{\max}-m_t}].
\end{equation}
where $\mathbf{e}_t \in \mathbb{R}^{d}$ denotes the task embedding for the $t$-th task, which identifies the task associated with the current training sample, $\mathbf{\lambda(\theta)}$ denotes the preference vector generated by the preference learning mechanism, and $\mathbf{0}_{d_{\max}-m_t}$ denotes the zero-padding vector of dimension $d_{\max}-m_t$, where $d_{\max} = \max\{m_1, m_2, \ldots, m_T\}$ denotes the maximum number of objectives among all tasks and $m_t$ denotes the number of objectives associated with the $t$-th task.
	
This design enables a unified neural network to simultaneously handle heterogeneous multi-objective optimization problems, achieving effective task distinction and parameter sharing through the task embedding. Furthermore, combining fixed task embedding with the random task scheduling in Algorithm~\ref{alg:proposed} mitigates catastrophic forgetting~\cite{Kirkpatrick2016OvercomingCF}. The former suppresses inter-task gradient interference at the input level, while the latter ensures balanced parameter updates across all tasks throughout training.

\textbf{Multitask Collaborative Training}: The pseudocode for multitask collaborative training is presented in Algorithm~\ref{alg:proposed}. The Transformer-based Pareto neural model $\mathbf{x}_{\beta}(\cdot)$ is initialized together with the optimizer $\mathcal{O}$ before training begins. At each iteration, the following steps are performed until the maximum number of iterations $n$ is reached. 

First, a task index $t$ is randomly selected from $\{1, 2, \ldots, T\}$ to identify the corresponding MOO task $\mathbf{P}_t$. Subsequently, $B$ preference vectors $\lambda(\theta)$ are uniformly sampled with $\theta \in [0, \pi/2]^{m-1}$. After selecting the task and sampling preferences, the task embedding $\mathbf{e}_t$ is concatenated with the preference vectors and a zero-padding vector to construct the network input $\mathbf{V}$. The input is then fed into the Pareto neural model $\mathbf{x}_{\beta}(\cdot)$ to generate the corresponding decision variables $\mathbf{X}_i$. Finally, these generated solutions are evaluated using the objective functions of task $\mathbf{P}_t$ to obtain their objective values $\mathbf{J}_i$.

Based on $\mathbf{X}_i$ and $\mathbf{J}_i$, the loss function for PSL is computed. To effectively train the model $\mathbf{x}_{\beta}(\cdot)$, we adopt the hypervolume maximization-based loss function PSL-HV1~\cite{Zhang2023HypervolumeMA}. The core idea of this loss function is to optimize model parameters by maximizing a surrogate form of the HV. The loss function is defined as follows:
\begin{equation}
    \label{eq:hv_loss}
	\begin{cases}
		\overline{\mathcal{H}}_r(\beta) = c_m \mathbb{E}_{\theta}[\rho_{\beta}(\theta)], \\[6pt]
		\rho_{\beta}(\theta) = \begin{cases}
		\rho(x_{\beta}(\theta), \theta)^m & \text{if } \rho(x_{\beta}(\theta), \theta) \geq 0 \\ 
		\rho(x_{\beta}(\theta), \theta) & \text{otherwise}
		\end{cases}
	\end{cases}.
\end{equation}
where $\rho(x_{\beta}(\theta), \theta)$ denotes the projected distance at angle $\theta$, satisfying $\rho(x_{\beta}(\theta), \theta) \leq \rho_{\mathcal{X}}(\theta)$ as defined in Eq.~(\ref{eq:f}). When the model approximates the entire PS, $\rho(x_{\beta}(\theta), \theta) \to \rho_{\mathcal{X}}(\theta)$ for all $\theta \in \Theta$, and consequently $\overline{\mathcal{H}}_r(\beta) \to \mathcal{H}_r(\mathcal{T})$, where $\mathcal{T}$ denotes the true PF. The normalization constant $c_m = \frac{\pi^{m/2}}{2^m \,\Gamma(m/2+1)}$ depends on the number of objectives $m$, and $\Gamma(\cdot)$ is the Gamma function.
	\begin{equation}
        \label{eq:f}
		\rho_\mathcal{X}(\theta) = \max_{x \in \mathcal{X}} \rho(x, \theta) = \max_{x \in \mathcal{X}} \min_{i \in [m]} \left\{ \frac{r_i - f_i(x)}{\lambda_i(\theta)} \right\}.
	\end{equation}
where $\mathbf{r} = (r_1, \dots, r_m) \in \mathbb{R}^m$ is a pre-specified reference point that dominates all solutions in the objective space, $r_i$ is its $i$-th coordinate, $f_i(x)$ is the $i$-th objective function value, and $\lambda_i(\theta)$ is the $i$-th component of the preference vector. 

\begin{algorithm}[h]
    \caption{Multitask Collaborative Training}
    \label{alg:proposed}
    \KwIn{%
    
        1) The number of tasks: $T$\;
        2) The set of multi-objective problems: 
            $\mathcal{P} = \{\mathbf{P}_1, \mathbf{P}_2, \ldots, \mathbf{P}_T\}$\;
        3) The number of training iterations: $n$\;
        4) The size of batch size: $B$\;
        5) The length of task embedding: $d$.
    }
    \KwOut{Trained Pareto neural network model $\mathbf{x}_{\beta}(\cdot)$, loss array $\mathit{Loss}$, and model quality evaluation $Q$.}

    Initialize the transformer model with labels of $\mathbf{x}_{\beta}(\cdot)$ and the optimizer with labels of $\mathcal{O}$. Set iteration $i \leftarrow 0$.

    \While{$i < n$}{
        Randomly select a task index $t$ from $\{1, 2, \ldots, T\}$ and retrieve task $\mathbf{P}_t$.\

        Uniformly sample $B$ preference vectors $\lambda(\theta)$ with $\theta \in [0,\, \pi/2]^{m-1}$.\

        \If{\upshape use extreme preferences}{
            $\lambda(\theta) \leftarrow \lambda(\theta) \cup 
            \{e_1, e_2, \ldots, e_m\}$\;
            \tcp{$e_j$ denotes the $j$-th standard basis vector}
        }

        Truncate $\lambda(\theta)$, each objective weight $\lambda_j(\theta)$ is restricted to the range $[0.01, 0.99]$.\

        Construct input: $\mathbf{V} = [\mathbf{e}_t, \boldsymbol{\lambda}(\theta), \mathbf{0}_{d_{\max}-m_t}]$.\

        $\mathbf{X}_i \leftarrow$ Forward pass through model $\mathbf{x}_{\beta}(\cdot)$ to obtain solutions.\

        $\mathbf{J}_i \leftarrow$ Evaluate solutions $\mathbf{X}_i$ using the objective functions of $\mathbf{P}_t$.\

        Compute PSL-HV1 loss $L$ based on $\mathbf{X}_i$ and $\mathbf{J}_i$ via Eq.~(\ref{eq:hv_loss}).\

        $\mathcal{O}$.zero\_grad(): $\nabla_{\beta}L \leftarrow 0$.\
        
        Backpropagate: $\nabla_{\beta}L \leftarrow \partial L / \partial \beta$.\
        
        $\mathit{Loss}[i] \leftarrow \texttt{detach}(L)$.\

        \If{gradient clipping is enabled}{
				Gradient clipping on model parameters\;
			}

        Update model parameters: 
        $\beta \leftarrow \mathcal{O}.\texttt{step}(\beta,\, \nabla_{\beta}L)$.\

        $i \leftarrow i + 1$.\
    }

    Compute model quality $Q$.\

    \Return{trained model $\mathbf{x}_{\beta}(\cdot)$, computed loss array $Loss$, and model quality $Q$.}
\end{algorithm}

After computing the loss, gradient clipping is applied to the model parameters to prevent gradient explosion and enhance training stability. The optimizer then updates the model parameters, completing one training iteration. Upon completion of training, the quality of the model is evaluated by computing a performance metric $Q$, which assesses the overall effectiveness of the model in approximating the PS.

\section{Experiments} \label{sec:Experiments}
In this section, we present comprehensive empirical evaluations conducted on multitask test suites that include both benchmark problems and real-world engineering applications. Algorithm performance is assessed using three metrics, namely hypervolume, range, and sparsity. We further conduct an ablation study to validate the architectural choice of the Transformer encoder backbone, an extended evaluation on the bbob-biobj test suite to assess generalizability, and a Wilcoxon signed-rank statistical analysis to confirm the significance of observed performance differences.

\subsection{Testing Problems}
We evaluate the proposed CoAction framework on a test suite comprising seven multi-objective optimization problems. Specifically, the suite includes four well-established benchmark problems including ZDT1-2 ($m=2$) and VLMOP1-2 ($m=2$), as well as three real-world engineering design problems namely Four Bar Truss Design (RE21, $m=2$), Hatch Cover Design (RE24, $m=2$), and Rocket Injector Design (RE37, $m=3$). The ZDT and VLMOP series possess analytically known PFs and exhibit diverse characteristics including both convex and non-convex PFs, which facilitates the assessment of algorithmic convergence and solution diversity. The engineering design problems span structural mechanics, mechanical design, and aerospace applications, providing a rigorous evaluation of optimization performance under realistic constraints. All objective values are normalized to $[0,1]$ to ensure fair comparisons across problems.
	
\subsection{Neural Model Architecture and Feasibility Guarantees}
We construct our Pareto neural model $\mathbf{x}_{\beta}(\cdot)$ based on a Transformer encoder backbone comprising the following components: (1) A linear projection layer that maps input sequences into 128-dimensional embeddings; (2) A positional encoding module that incorporates 128-dimensional positional information~\cite{Baniata2022ARP} with a dropout rate of 0.1; (3) A single-layer encoder block consisting of 4 attention heads and a two-layer feed-forward network with a hidden dimension of 128, employing Pre-Norm layer normalization~\cite{Xiong2020OnLN} with residual connections and a dropout rate of 0.1; (4) A weighted pooling module that computes attention weights via a two-layer MLP with a hidden dimension of 64 and Tanh activation, followed by softmax normalization and weighted summation to produce a global sequence representation; (5) Task-specific output heads that map the global representation to the decision variables of each task.

The network is optimized using the AdamW-ScheduleFree algorithm~\cite{Defazio2024TheRL} with a learning rate of $2 \times 10^{-3}$, which requires no learning rate warmup, with a batch size of 256. For constrained problems, a sigmoid activation followed by a linear rescaling is applied to the output layer, mapping predictions into the feasible range $[l, u]$ via $\hat{x} = l + (u - l) \cdot \sigma(z)$, where $l$ and $u$ denote the lower and upper bounds of the decision variables, respectively. For unconstrained problems, a linear output layer is used to directly produce the decision variables.

\subsection{Metrics and Results}
To evaluate the quality of the learned Pareto solutions, we employ three indicators: (1) the HV indicator, which measures the volume of objective space dominated by the solution set, with higher values indicating better coverage; (2) the Range indicator, which quantifies the spread of the PF, with higher values indicating a broader solution distribution; and (3) the Sparsity metric, where lower values are preferable, indicating more evenly distributed solutions.

Table~\ref{tab:comparison_results1} compares the proposed CoAction framework against the single-task PSL-HV1 baseline across all seven multi-objective optimization problems. We analyze the results from three perspectives: stability, solution quality, and computational efficiency.

\begin{table*}[htbp]
\centering 
\caption{Comparison of PSL results between the proposed CoAction framework and the single-task baseline on seven multi-objective optimization problems. Results are reported as mean $\pm$ standard deviation over five independent runs. \textbf{Bold} values indicate better performance.}
\label{tab:comparison_results1}
\renewcommand{\arraystretch}{1.15}
\setlength{\tabcolsep}{6pt}
\resizebox{\textwidth}{!}{%
\begin{tabular}{l|ccc|ccc}
\toprule
\multirow{2}{*}{\textbf{Method}} 
  & \multicolumn{3}{c|}{\textbf{Single-task PSL-HV1}} 
  & \multicolumn{3}{c}{\textbf{Multitask PSL-HV1 (Ours)}} \\
\cmidrule(lr){2-4}\cmidrule(lr){5-7}
 & HV$\uparrow$ & Range$\uparrow$ & Sparsity$\downarrow$ 
 & HV$\uparrow$ & Range$\uparrow$ & Sparsity$\downarrow$ \\
\midrule
ZDT1   & $11.79_{\pm0.0075}$ & $1.52_{\pm0.0040}$ & $0.73_{\pm0.0110}$ & $\mathbf{11.90_{\pm0.0000}}$ & $\mathbf{1.57_{\pm0.0000}}$ & $0.85_{\pm0.0009}$ \\
ZDT2   & $11.29_{\pm0.0080}$ & $1.46_{\pm0.0000}$ & $0.73_{\pm0.0080}$ & $\mathbf{11.56_{\pm0.0001}}$ & $\mathbf{1.57_{\pm0.0000}}$ & $1.09_{\pm0.0027}$ \\
VLMOP1 & $\mathbf{12.07_{\pm0.0000}}$ & $\mathbf{1.57_{\pm0.0000}}$ & $0.90_{\pm0.1160}$ & $12.00_{\pm0.0027}$ & $1.56_{\pm0.0005}$ & $1.18_{\pm0.0399}$ \\
VLMOP2 & $\mathbf{11.57_{\pm0.0050}}$ & $\mathbf{1.56_{\pm0.0000}}$ & $0.98_{\pm0.0260}$ & $11.39_{\pm0.0051}$ & $1.50_{\pm0.0006}$ & $1.20_{\pm0.0165}$ \\
RE21   & $11.90_{\pm0.0050}$ & $1.56_{\pm0.0050}$ & $0.84_{\pm0.0380}$ & $\mathbf{12.08_{\pm0.0000}}$ & $\mathbf{1.57_{\pm0.0000}}$ & $0.56_{\pm0.0003}$ \\
RE24   & $12.17_{\pm0.0280}$ & $1.52_{\pm0.0240}$ & $0.63_{\pm0.2780}$ & $\mathbf{12.20_{\pm0.0000}}$ & $\mathbf{1.57_{\pm0.0000}}$ & $1.54_{\pm0.0349}$ \\
RE37   & $\mathbf{40.84_{\pm0.0110}}$ & $0.69_{\pm0.0120}$ & $3.80_{\pm0.5010}$ & $40.62_{\pm0.0541}$ & $\mathbf{0.87_{\pm0.0259}}$ & $6.96_{\pm3.8894}$ \\
\midrule
\textit{Mean}
  & $15.95$ & $1.41$ & $1.23$
  & $\mathbf{15.96}$ & $\mathbf{1.46}$ & $1.91$ \\
\midrule
\textit{Time(s) $\downarrow$}
  & \multicolumn{3}{c|}{$370.25$}
  & \multicolumn{3}{c}{$\mathbf{270.13}$ \ (\textbf{$-$27\%})} \\
\bottomrule
\end{tabular}%
}
\end{table*}

\begin{figure*}[htbp]
\centering
\begin{subfigure}[b]{0.24\linewidth}
    \centering
    \includegraphics[width=\textwidth]{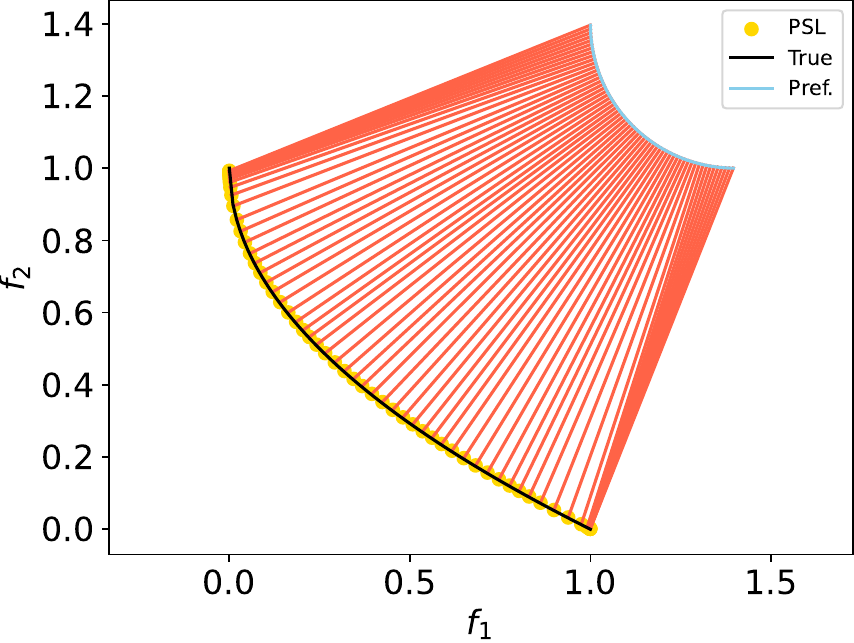}
    \caption{ZDT1\textsubscript{TF}}
    \label{fig:zdt1_transformer}
\end{subfigure}
\hfill
\begin{subfigure}[b]{0.24\linewidth}
    \centering
    \includegraphics[width=\textwidth]{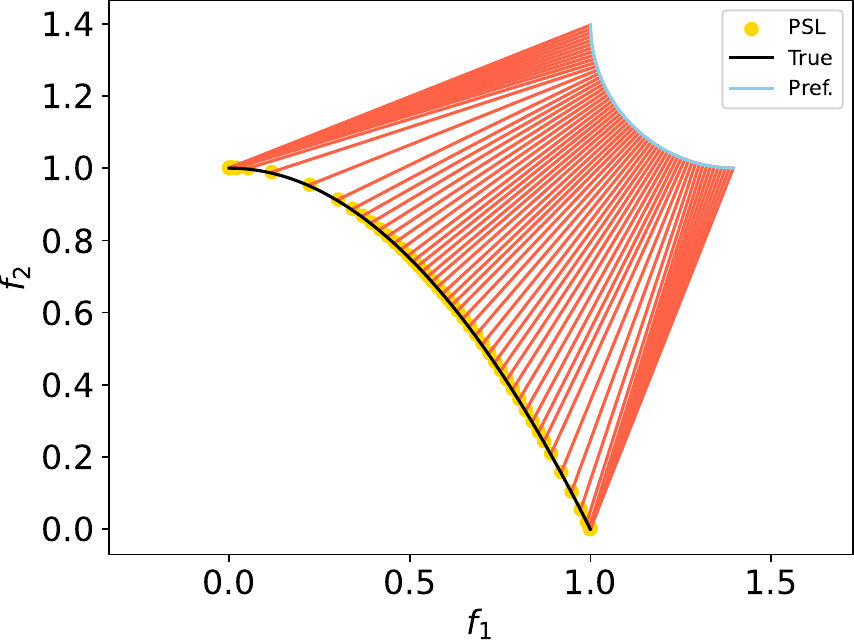}
    \caption{ZDT2\textsubscript{TF}}
    \label{fig:zdt2_transformer}
\end{subfigure}
\hfill
\begin{subfigure}[b]{0.24\linewidth}
    \centering
    \includegraphics[width=\textwidth]{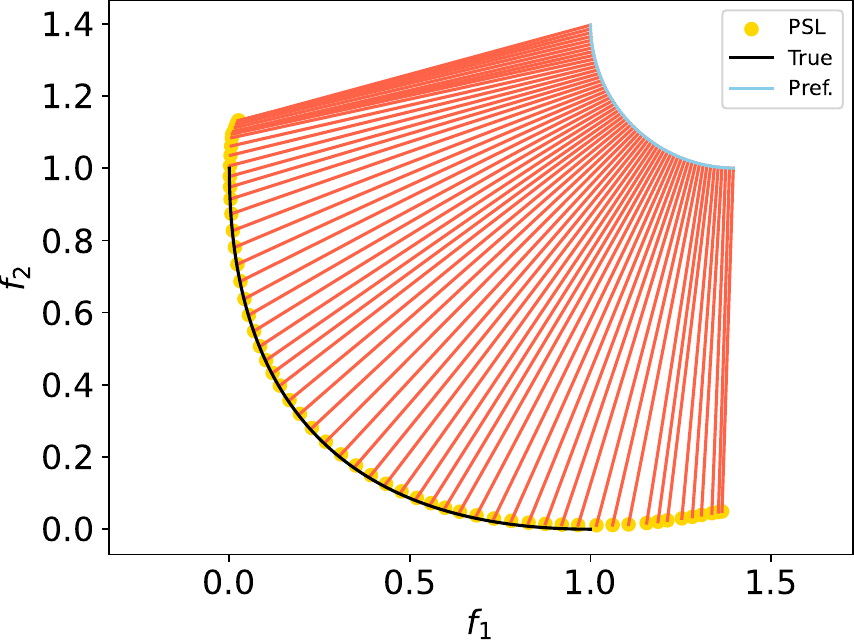}
    \caption{VLMOP1\textsubscript{TF}}
    \label{fig:vlmop1_transformer}
\end{subfigure}
\hfill
\begin{subfigure}[b]{0.24\linewidth}
    \centering
    \includegraphics[width=\textwidth]{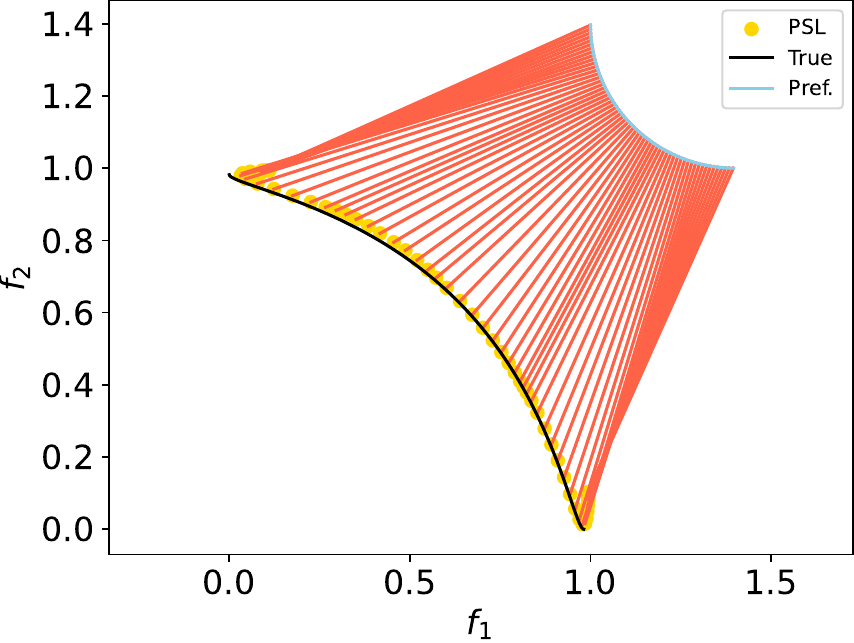}
    \caption{VLMOP2\textsubscript{TF}}
    \label{fig:vlmop2_transformer}
\end{subfigure}
\vspace{0.1cm}

\begin{subfigure}[b]{0.24\linewidth}
    \centering
    \includegraphics[width=\textwidth]{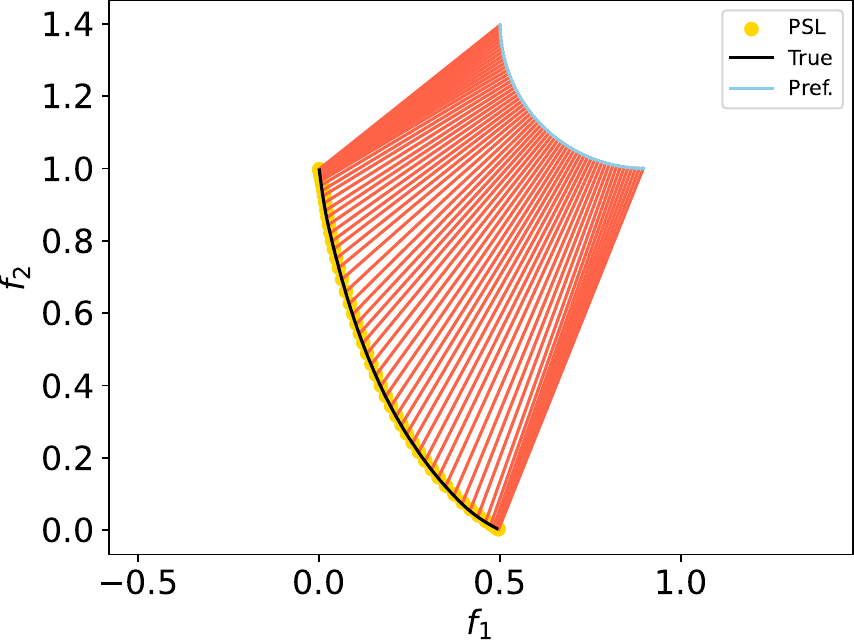}
    \caption{RE21\textsubscript{TF}}
    \label{fig:re21_transformer}
\end{subfigure}
\hfill
\begin{subfigure}[b]{0.24\linewidth}
    \centering
    \includegraphics[width=\textwidth]{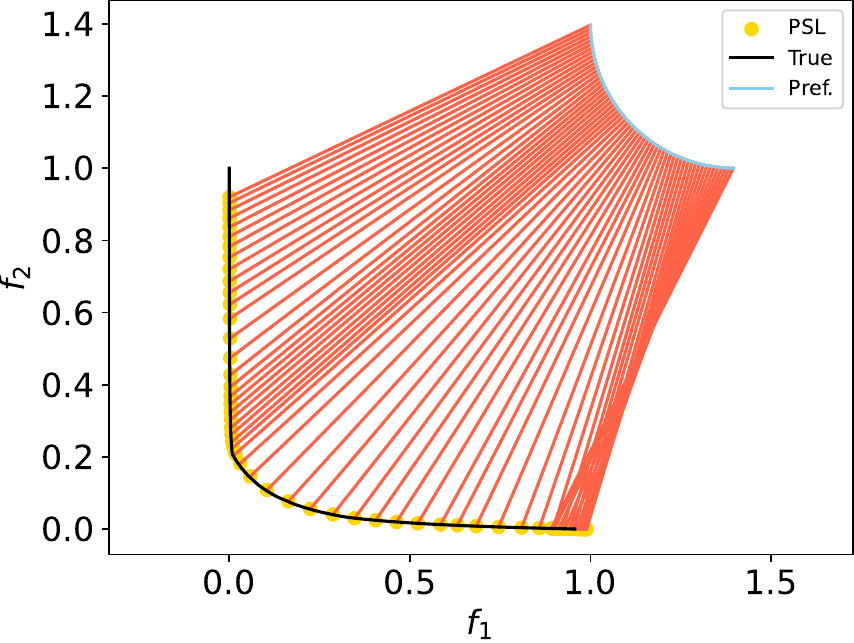}
    \caption{RE24\textsubscript{TF}}
    \label{fig:re24_transformer}
\end{subfigure}
\hfill
\begin{subfigure}[b]{0.24\linewidth}
    \centering
    \includegraphics[width=\textwidth]{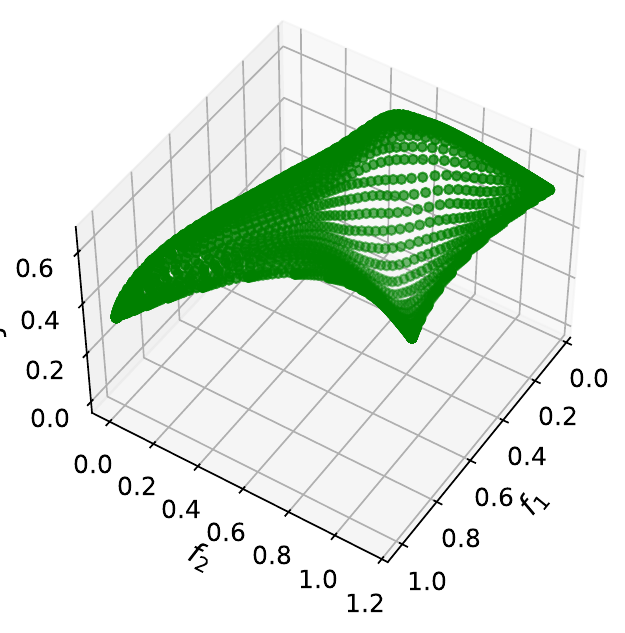}
    \caption{RE37\textsubscript{TF}}
    \label{fig:re37_transformer}
\end{subfigure}
\vspace{0.1cm}

\caption{Learned Pareto solutions with the Transformer (TF) backbone on seven multiobjective multitask problems. The colored surfaces in (a)--(g) represent the Pareto solution distributions obtained by the CoAction framework.}
\label{fig:TF_results}
\end{figure*}

\noindent\textbf{Stability.} Variance is consistently low across most problems, with HV standard deviations approaching zero on ZDT1, ZDT2, RE21, and RE24, confirming stable convergence. The exception is RE37, where Sparsity exhibits higher variability ($\pm$3.8894). This tri-objective problem features a more irregular PF, making it harder for the shared backbone to generate uniformly distributed solutions. Notably, single-task training on RE37 also yields non-trivial Sparsity variance ($\pm$0.5010), confirming that this instability is intrinsic to the problem structure rather than a consequence of the multitask architecture.

\noindent\textbf{Solution quality.} The CoAction framework achieves higher HV and Range on ZDT1, ZDT2, RE21, and RE24, demonstrating a clear benefit from cross-task knowledge sharing. On RE21 in particular, the multitask setting simultaneously improves all three metrics (HV: 12.08 vs.\ 11.90, Range: 1.57 vs.\ 1.56, Sparsity: 0.56 vs.\ 0.84), indicating that shared representations not only broaden front coverage but also promote more uniform solution distribution. On VLMOP1, VLMOP2, and RE37, single-task training yields marginally higher HV, indicating mild negative transfer. This is attributed to geometric dissimilarity among PFs: VLMOP1 and VLMOP2 have convex symmetric fronts, whereas ZDT1, ZDT2, and RE problems feature curved or irregular geometries, leading to conflicting gradient signals during joint optimization. Nevertheless, all HV differences remain within 1\% across problems, confirming that solution quality is well preserved overall. The higher Sparsity values observed under CoAction framework are consistent with its improved Range. Since a broader learned front naturally increases average inter-solution distances, elevated Sparsity should be interpreted as a consequence of better front coverage rather than degraded solution uniformity~\cite{Zhang2023HypervolumeMA}.

To provide an overall view, we report the mean of each metric across all seven problems. The CoAction framework achieves a higher mean HV ($15.96$ vs.\ $15.95$) and a higher mean Range ($1.46$ vs.\ $1.41$), confirming that multitask training consistently broadens PF coverage on average. The mean Sparsity under CoAction framework is higher ($1.91$ vs.\ $1.23$), which is largely attributable to the elevated Sparsity on RE37. As discussed, this reflects improved front spread rather than degraded solution uniformity.

\noindent\textbf{Computational efficiency.} The CoAction framework trains all seven tasks jointly in 5000 iterations, while the single-task baseline requires 7 $\times$ 1000 = 7000 iterations. Despite this reduction in total iterations, the proposed CoAction framework achieves competitive solution quality with a 27\% reduction in computational time (370.25s vs.\ 270.13s).

Together, the near-zero HV standard deviations and consistent performance confirm that catastrophic forgetting is not significant. This robustness stems from the architecture. Although the shared backbone receives gradients from all seven tasks, each task retains a dedicated output head for task-specific adjustments. Consequently, task-specific representations are not directly overwritten by updates from other tasks, reducing interference and stabilizing convergence.

Fig.~\ref{fig:TF_results} shows the distribution of Pareto solutions learned by the PSL-HV1 algorithm across all seven test problems. As shown in Table~\ref{tab:comparison_results1} and Fig.~\ref{fig:TF_results}, the proposed method consistently produces well-distributed solution sets that closely approximate the true PFs. These results demonstrate strong convergence behavior and solution diversity across both benchmark problems and real-world engineering applications. In particular, the strong performance on the challenging tri-objective RE37 problem provides initial evidence of the scalability and generalization of the Transformer-based backbone.

\subsection{Ablation Study: Backbone Architecture}
To validate the architectural choice of a Transformer encoder, we compare it against a standard MLP backbone under identical training settings, varying only the backbone architecture.

\begin{table*}[htbp]
\centering 
\caption{Comparison of PSL results between the Transformer backbone and the MLP backbone on seven multi-objective optimization problems. Results are reported as mean $\pm$ standard deviation over five independent runs. \textbf{Bold} values indicate better performance.}
\label{tab:comparison_results2}
\renewcommand{\arraystretch}{1.15}
\setlength{\tabcolsep}{6pt}
\resizebox{\textwidth}{!}{%
\begin{tabular}{l|ccc|ccc}
\toprule
\multirow{2}{*}{\textbf{Method}} 
  & \multicolumn{3}{c|}{\textbf{Transformer backbone (ours)}} 
  & \multicolumn{3}{c}{\textbf{MLP backbone}} \\
\cmidrule(lr){2-4}\cmidrule(lr){5-7}
 & HV$\uparrow$ & Range$\uparrow$ & Sparsity$\downarrow$ 
 & HV$\uparrow$ & Range$\uparrow$ & Sparsity$\downarrow$ \\
\midrule
ZDT1   
  & $\mathbf{11.90_{\pm0.0000}}$ & $\mathbf{1.57_{\pm0.0000}}$ & $0.85_{\pm0.0009}$
  & $11.85_{\pm0.0324}$ & $1.55_{\pm0.0134}$ & $0.74_{\pm0.0409}$ \\
ZDT2   
  & $\mathbf{11.56_{\pm0.0001}}$ & $\mathbf{1.57_{\pm0.0000}}$ & $1.09_{\pm0.0027}$
  & $10.59_{\pm0.9272}$ & $1.09_{\pm0.4807}$ & $0.75_{\pm0.5400}$ \\
VLMOP1 
  & $12.00_{\pm0.0027}$ & $\mathbf{1.56_{\pm0.0005}}$ & $1.18_{\pm0.0399}$
  & $\mathbf{12.03_{\pm0.0241}}$ & $1.56_{\pm0.0114}$ & $0.86_{\pm0.1314}$ \\
VLMOP2 
  & $11.39_{\pm0.0051}$ & $\mathbf{1.50_{\pm0.0006}}$ & $1.20_{\pm0.0165}$
  & $\mathbf{11.41_{\pm0.1123}}$ & $1.50_{\pm0.0458}$ & $0.81_{\pm0.0812}$ \\
RE21   
  & $\mathbf{12.08_{\pm0.0000}}$ & $\mathbf{1.57_{\pm0.0000}}$ & $0.56_{\pm0.0003}$
  & $12.07_{\pm0.0055}$ & $1.57_{\pm0.0055}$ & $0.60_{\pm0.0152}$ \\
RE24   
  & $\mathbf{12.20_{\pm0.0000}}$ & $\mathbf{1.57_{\pm0.0000}}$ & $1.54_{\pm0.0349}$
  & $11.97_{\pm0.2919}$ & $1.08_{\pm0.6699}$ & $0.44_{\pm0.5110}$ \\
RE37
  & $\mathbf{40.62_{\pm0.0541}}$ & $\mathbf{0.87_{\pm0.0259}}$ & $6.96_{\pm3.8894}$
  & $36.07_{\pm1.4546}$ & $0.21_{\pm0.0130}$ & $1.13_{\pm0.8205}$ \\
\midrule
\textit{Mean}
  & $\mathbf{15.96}$ & $\mathbf{1.46}$ & $1.91$
  & $15.14$ & $1.22$ & $0.76$ \\
\bottomrule
\end{tabular}%
}
\end{table*}

\begin{figure*}[h]
\centering
\begin{subfigure}[b]{0.24\linewidth}
    \centering
    \includegraphics[width=\textwidth]{ZDT1.pdf}
    \caption{ZDT1\textsubscript{TF}}
    \label{fig:zdt1_transformer}
\end{subfigure}
\hfill
\begin{subfigure}[b]{0.24\linewidth}
    \centering
    \includegraphics[width=\textwidth]{VLMOP1.pdf}
    \caption{VLMOP1\textsubscript{TF}}
    \label{fig:vlmop1_transformer}
\end{subfigure}
\hfill
\begin{subfigure}[b]{0.24\linewidth}
    \centering
    \includegraphics[width=\textwidth]{RE24.pdf}
    \caption{RE24\textsubscript{TF}}
    \label{fig:re24_transformer}
\end{subfigure}
\hfill
\begin{subfigure}[b]{0.24\linewidth}
    \centering
    \includegraphics[width=\textwidth]{RE37.pdf}
    \caption{RE37\textsubscript{TF}}
    \label{fig:re37_transformer}
\end{subfigure}
\vspace{0.1cm}

\begin{subfigure}[b]{0.24\linewidth}
    \centering
    \includegraphics[width=\textwidth]{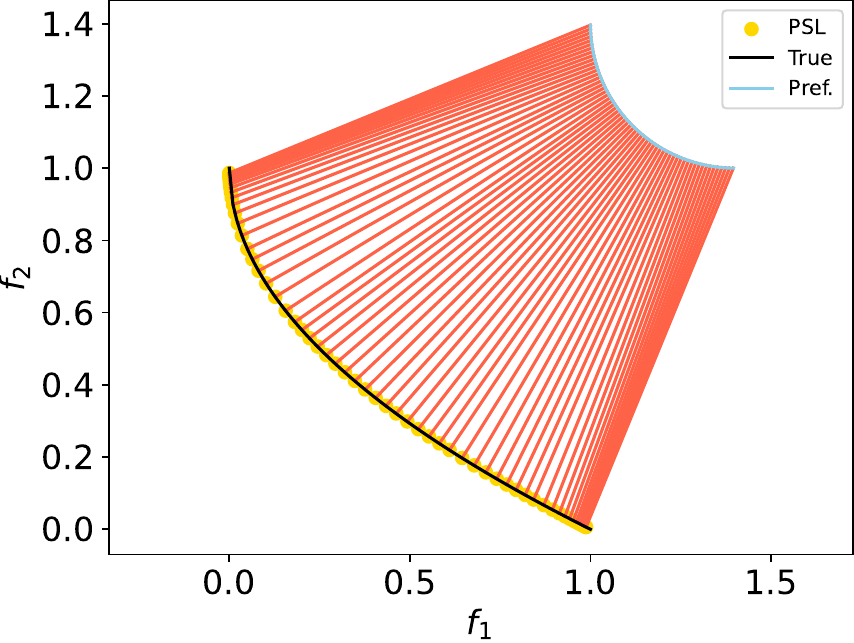}
    \caption{ZDT1\textsubscript{MLP}}
    \label{fig:zdt1_mlp}
\end{subfigure}
\hfill
\begin{subfigure}[b]{0.24\linewidth}
    \centering
    \includegraphics[width=\textwidth]{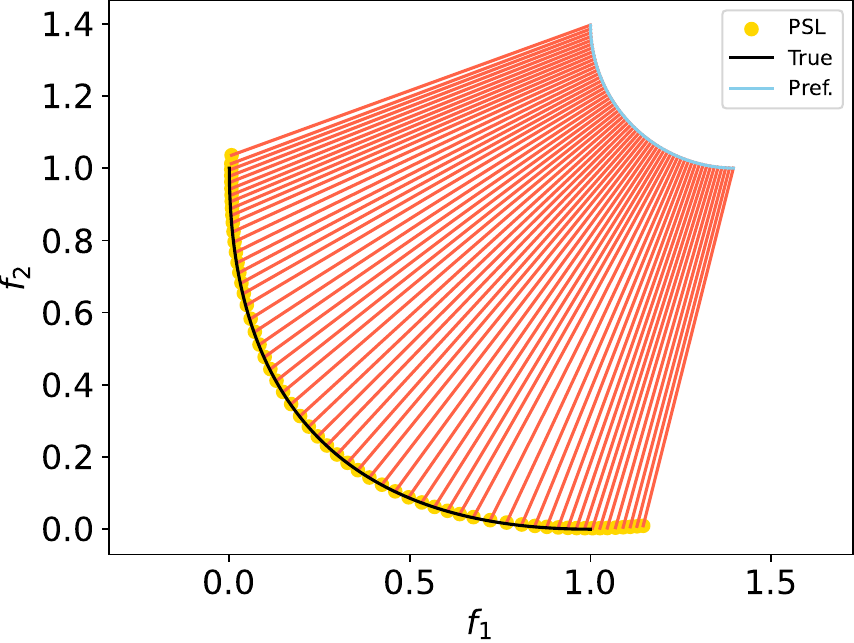}
    \caption{VLMOP1\textsubscript{MLP}}
    \label{fig:vlmop1_mlp}
\end{subfigure}
\hfill
\begin{subfigure}[b]{0.24\linewidth}
    \centering
    \includegraphics[width=\textwidth]{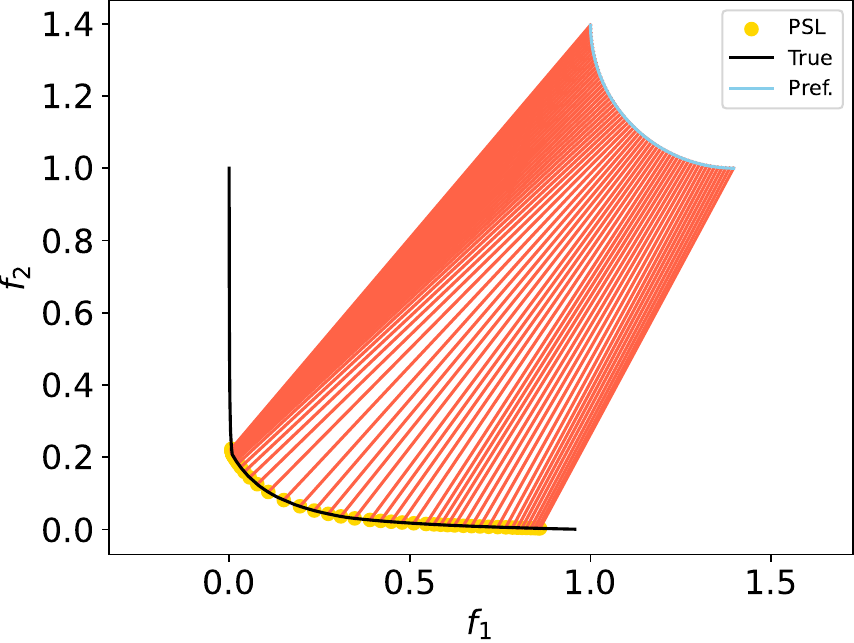}
    \caption{RE24\textsubscript{MLP}}
    \label{fig:re24_mlp}
\end{subfigure}
\hfill
\begin{subfigure}[b]{0.24\linewidth}
    \centering
    \includegraphics[width=\textwidth]{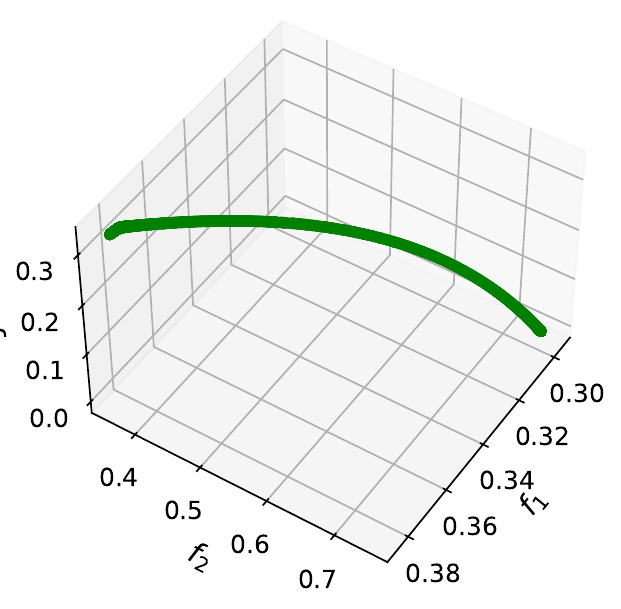}
    \caption{RE37\textsubscript{MLP}}
    \label{fig:re37_mlp}
\end{subfigure}
\vspace{0.1cm}
\caption{Learned Pareto solutions for four problems (ZDT1, VLMOP1, RE24, and RE37) with the Transformer (TF) and MLP backbones. The colored surfaces represent the Pareto solution distributions learned by each model.}
\label{fig:comparison_results}
\end{figure*}

Table~\ref{tab:comparison_results2} compares the Transformer backbone against the MLP backbone across all seven multi-objective optimization problems, and Fig.~\ref{fig:comparison_results} visualizes the Pareto solutions learned by both models on four of these problems (ZDT1, VLMOP1, RE24, and RE37).

Compared with the Transformer backbone results in Table~\ref{tab:comparison_results2}, the MLP backbone achieves competitive performance on all bi-objective problems, with only marginal differences in HV and Range. A notable performance gap is observed for the tri-objective RE37 problem. The Transformer achieves an HV of 40.62 compared to 36.07 for the MLP ($\sim$12.6\% improvement), and a range of 0.87 compared to 0.21 ($\sim$314\% improvement). As shown in Fig.~\ref{fig:comparison_results}, the MLP concentrates its solutions within a narrow region of the objective space on RE37, whereas the Transformer produces a well-distributed approximation that closely follows the true PF geometry. The higher Sparsity values observed under the Transformer backbone are consistent with its improved Range across problems. Since a broader learned front naturally increases average inter-solution distances, elevated Sparsity should be interpreted as a consequence of better front coverage rather than degraded solution uniformity~\cite{Zhang2023HypervolumeMA}. Regarding stability, the Transformer exhibits substantially lower standard deviations across nearly all bi-objective benchmarks and more consistent HV on RE37, indicating more robust optimization behavior overall.

To provide an overall view, we report the mean of each metric across all seven problems. The Transformer backbone achieves a higher mean HV ($15.96$ vs.\ $15.14$) and a higher mean Range ($1.46$ vs.\ $1.22$), confirming that the self-attention mechanism consistently broadens PF coverage on average. The correspondingly higher mean Sparsity ($1.91$ vs.\ $0.76$) is consistent with this broader coverage, as discussed above.

We attribute the performance gap on RE37, as well as the overall improvement in HV and Range, to the self-attention mechanism of the Transformer, which captures global dependencies across all input positions simultaneously, enabling the robust optimization behavior noted above. This property becomes especially critical when the PF geometry grows more complex in higher-dimensional objective spaces.

\subsection{Extended Evaluation: bbob-biobj Test Suite}
To further assess the generalizability of the proposed CoAction framework, we conducted an extended evaluation of the Bi-objective Black Box Optimization Benchmarking (bbob-biobj) test suite~\cite{Tuar2016COCOTB, Brockhoff2022BBOB}. Specifically, we construct a multitask benchmark comprising five bi-objective problems, each sharing the BBOB function $f_1$ as one component while pairing it with five different functions: $f_1$, $f_2$, $f_3$, $f_4$, and $f_5$. Each task operates in a 10-dimensional decision space with ReLU activation, and the model is trained for 1500 epochs.

\begin{table*}[htbp]
\centering
\caption{Hypervolume results on the bbob-biobj test suite after 1500 epochs, which consists of five bi-objective tasks each pairing the BBOB function $f_1$ with $f_1$, $f_2$, $f_3$, $f_4$, or $f_5$ in a 10-dimensional decision space.}
\label{tab:bbob_results}
\renewcommand{\arraystretch}{1.15}
\begin{tabular*}{0.55\linewidth}{@{\extracolsep{\fill}} l r}
\toprule
\textbf{Task} & \textbf{HV} \\
\midrule
bbob-biobj ($f_1 + f_1$) & 3.8302 \\
bbob-biobj ($f_1 + f_2$) & 3.8705 \\
bbob-biobj ($f_1 + f_3$) & 3.8271 \\
bbob-biobj ($f_1 + f_4$) & 3.8203 \\
bbob-biobj ($f_1 + f_5$) & 3.7496 \\
\midrule
\textit{Mean} & \textbf{3.8196} \\
\bottomrule
\end{tabular*}
\end{table*}

Table~\ref{tab:bbob_results} reports the per-task HV after 1500 epochs. All five tasks achieve consistently high HV values in the range $[3.74, 3.88]$, with a mean of $3.8196$. Nevertheless, the narrow spread ($\Delta\mathrm{HV} < 0.13$) confirms that the CoAction framework generalizes robustly.

\begin{figure*}[htbp]
\centering
\begin{subfigure}[b]{0.32\linewidth}
    \includegraphics[width=\textwidth]{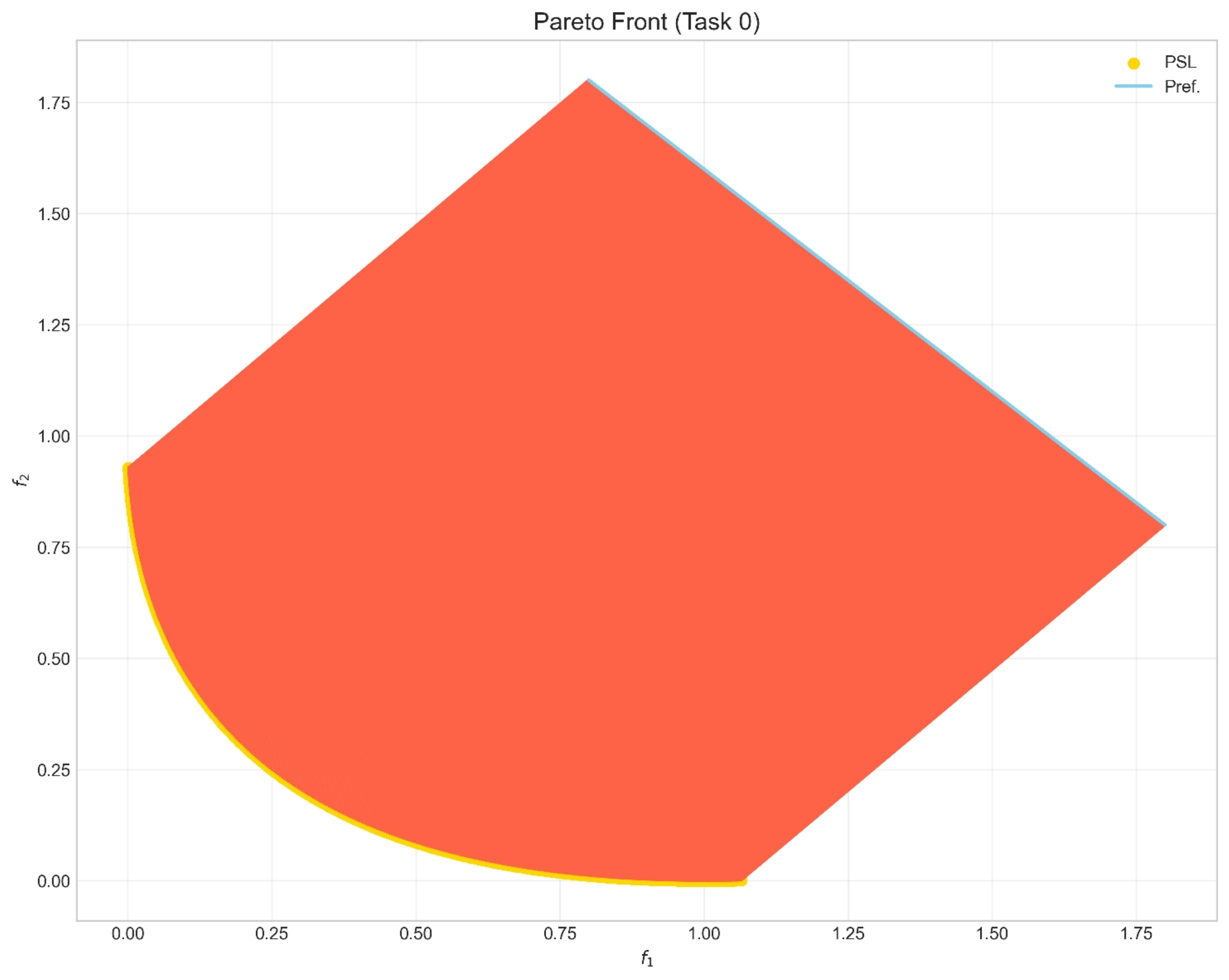}
    \caption{bbob-biobj ($f_1 + f_1$)}
    \label{fig:bbob_f1_f1}
\end{subfigure}
\hfill
\begin{subfigure}[b]{0.32\linewidth}
    \includegraphics[width=\textwidth]{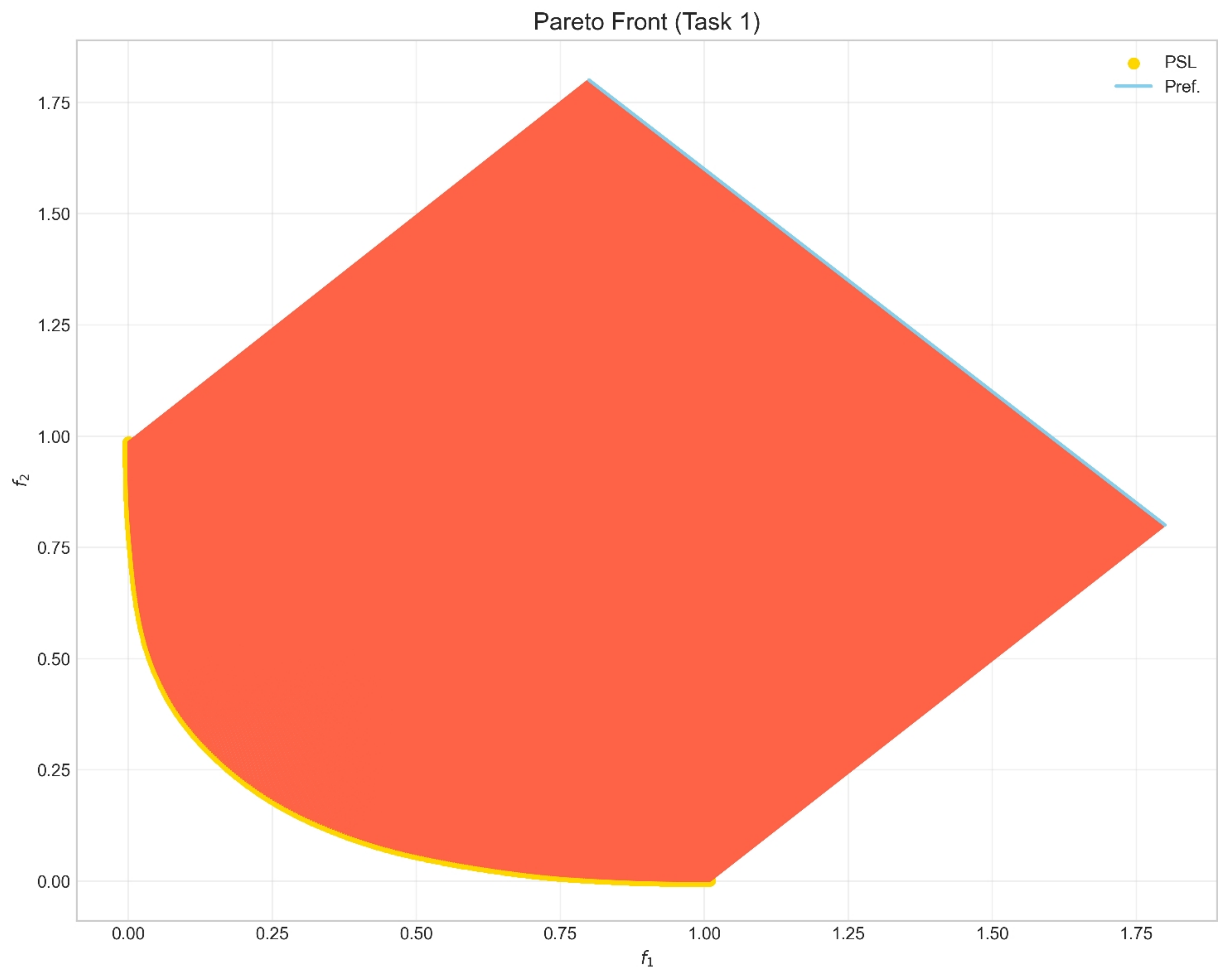}
    \caption{bbob-biobj ($f_1 + f_2$)}
    \label{fig:bbob_f1_f2}
\end{subfigure}
\hfill
\begin{subfigure}[b]{0.32\linewidth}
    \includegraphics[width=\textwidth]{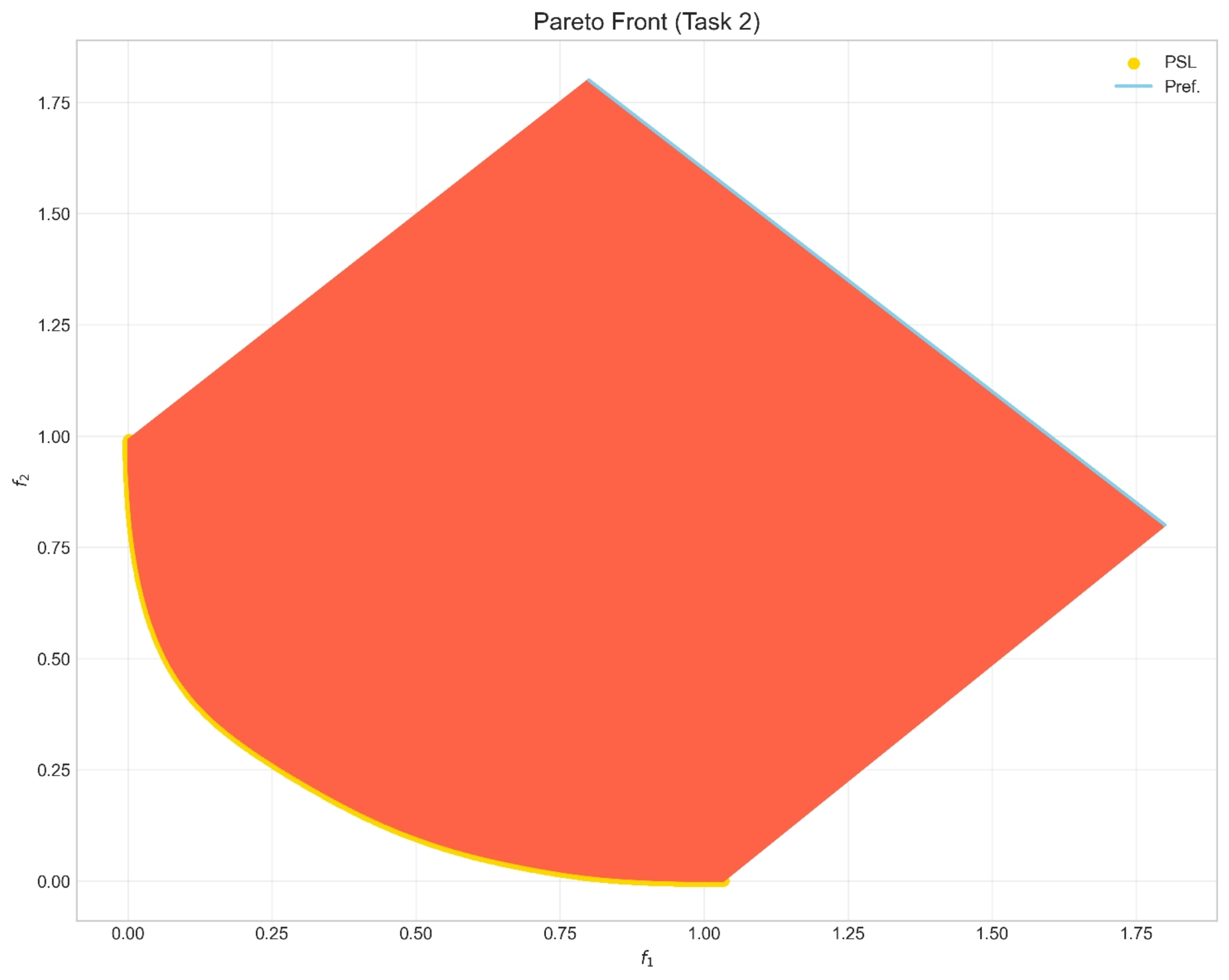}
    \caption{bbob-biobj ($f_1 + f_3$)}
    \label{fig:bbob_f1_f3}
\end{subfigure}

\vspace{0.1cm}
\centering
\begin{subfigure}[b]{0.32\linewidth}
    \includegraphics[width=\textwidth]{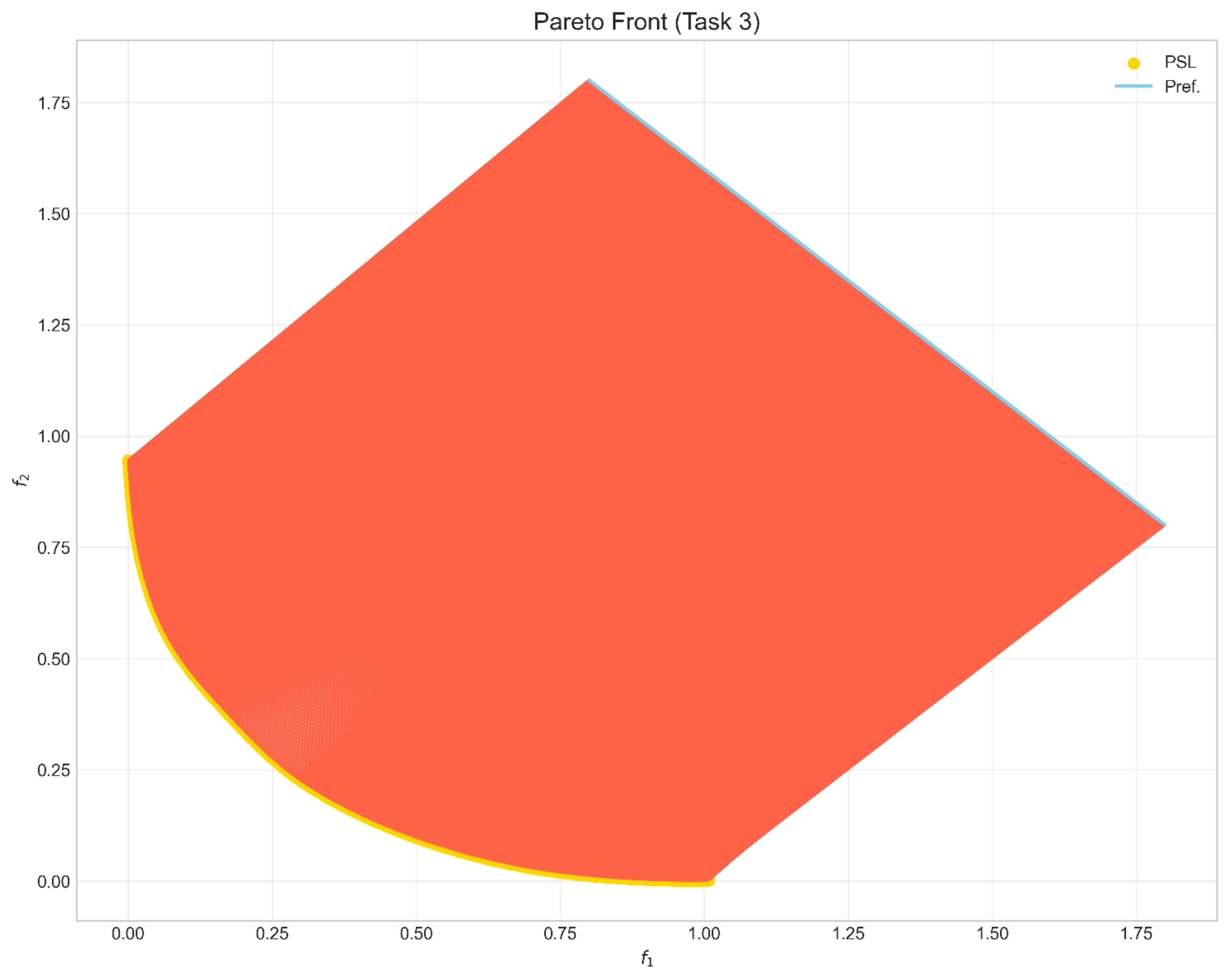}
    \caption{bbob-biobj ($f_1 + f_4$)}
    \label{fig:bbob_f1_f4}
\end{subfigure}
\hspace{0.02\linewidth}
\begin{subfigure}[b]{0.32\linewidth}
    \includegraphics[width=\textwidth]{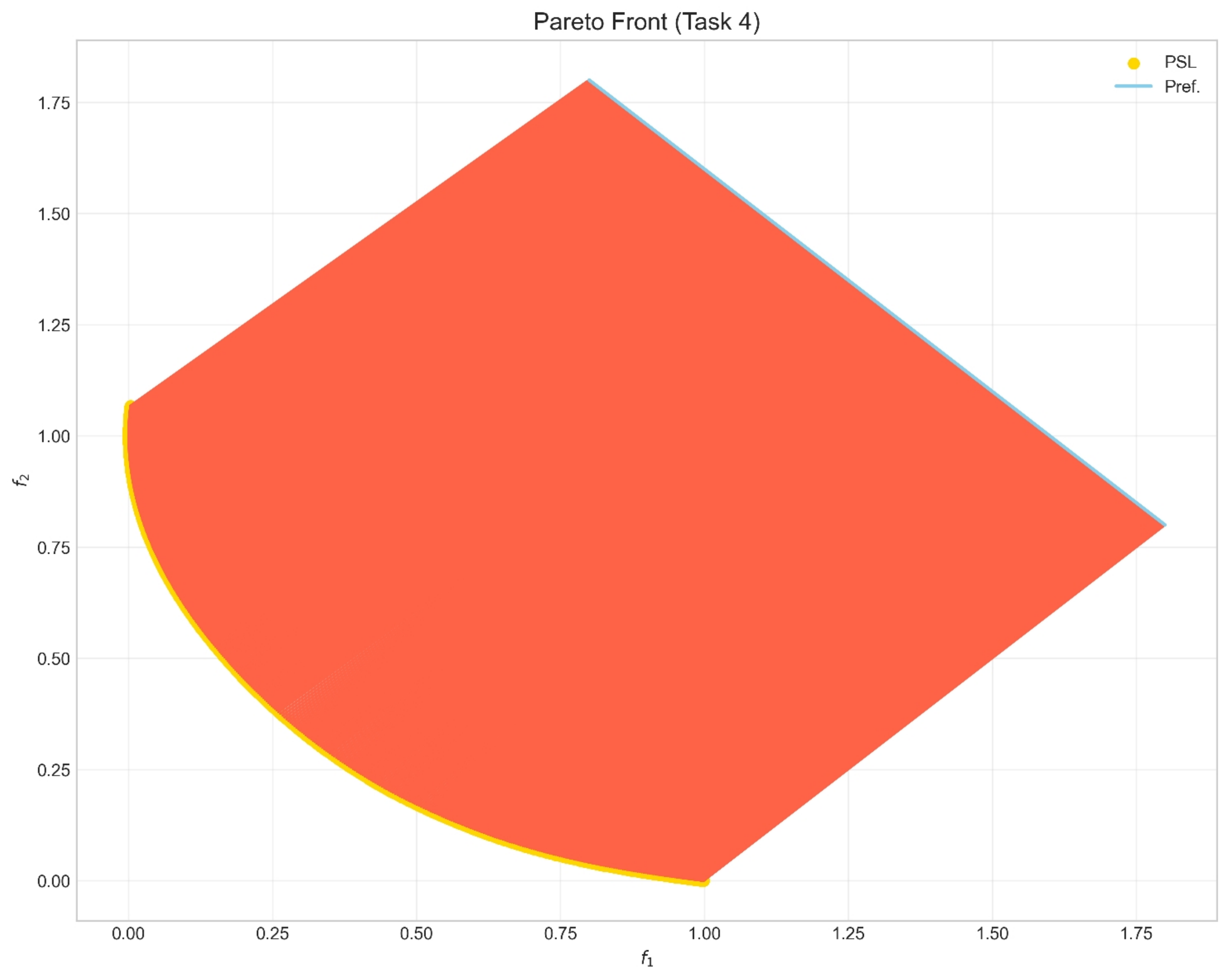}
    \caption{bbob-biobj ($f_1 + f_5$)}
    \label{fig:bbob_f1_f5}
\end{subfigure}
\caption{Learned Pareto solutions on five bbob-biobj problems after 1500 epochs. The orange surfaces represent the learned solutions and the yellow boundary denotes the PFs.}
\label{fig:bbob}
\end{figure*}

Fig.~\ref{fig:bbob} shows the learned solutions across all five tasks after epoch 1500. The learned solutions closely align with the PFs, demonstrating accurate approximation across diverse function pairings. 

\subsection{Statistical Analysis: Wilcoxon Signed-Rank Test}
To rigorously evaluate the statistical significance of performance differences, we employ the Wilcoxon signed-rank test~\cite{Wilcoxon1945IndividualCB} across all seven benchmark problems. Pairwise comparisons are conducted between (i) the proposed CoAction framework with Transformer backbone and the single-task baseline, and (ii) the Transformer backbone and the MLP backbone within the multitask framework. Each method is evaluated over five independent runs. Under this setting, the minimum achievable two-sided $p$-value under the Wilcoxon signed-rank test is $0.0625$. Given the small-sample constraint ($n=5$), we adopt a relaxed significance level of $\alpha$ = 0.10, following the recommendation of Derrac et al.~\cite{DERRAC20113} to balance Type I and Type II errors under limited experimental budgets.

\begin{table*}[ht]
\centering
\caption{Wilcoxon signed-rank test results: the proposed CoAction framework vs.\ Single-task baseline. $+:$ our method better, $=:$ equivalent means, $-:$ our method worse. Two-sided test, $n=5$, $\alpha=0.10$.}
\label{tab:wilcoxon_vs_single}
\begin{tabular*}{\textwidth}{@{\extracolsep{\fill}} l cc cc cc}
\toprule
\multirow{2}{*}{Problem}
  & \multicolumn{2}{c}{HV $\uparrow$}
  & \multicolumn{2}{c}{Range $\uparrow$}
  & \multicolumn{2}{c}{Sparsity $\downarrow$} \\
\cmidrule(lr){2-3}\cmidrule(lr){4-5}\cmidrule(lr){6-7}
  & $p$ & Dir. & $p$ & Dir. & $p$ & Dir. \\
\midrule
ZDT1      & 0.0625 & $+$ & 0.0625 & $+$ & 0.0625 & $-$ \\
ZDT2      & 0.0625 & $+$ & 0.0625 & $+$ & 0.0625 & $-$ \\
VLMOP1    & 0.0625 & $-$ & 0.1250 & $-$ & 0.0625 & $-$ \\
VLMOP2    & 0.0625 & $-$ & 0.0625 & $-$ & 0.0625 & $-$ \\
RE21      & 0.0625 & $+$ & 1.0000 & $+$ & 0.0625 & $+$ \\
RE24      & 0.0625 & $+$ & 0.0625 & $+$ & 0.0625 & $-$ \\
RE37      & 0.1875 & $-$ & 0.1875 & $+$ & 0.0625 & $-$ \\
\bottomrule
\end{tabular*}
\end{table*}

\textbf{CoAction Framework vs. Single-Task Baseline (Table~\ref{tab:wilcoxon_vs_single})}. Across 21 metric--problem combinations, the CoAction framework achieves statistically significant improvements in both HV and Range on ZDT1, ZDT2, and RE24 (all $p = 0.0625$), as well as in HV alone on RE21 ($p = 0.0625$), where Range shows no significant difference ($p = 1.0000$), confirming consistent cross-task benefit on these problems. On VLMOP2, mild negative transfer leads to statistically significant disadvantages in both HV and Range ($p = 0.0625$). On VLMOP1, only the HV disadvantage reaches significance ($p = 0.0625$), while the Range difference does not ($p = 0.1250$). On RE37, neither HV nor Range differences are statistically significant ($p = 0.1875$). Regarding Sparsity, the CoAction framework significantly reduces it on RE21 ($p = 0.0625$), whereas higher Sparsity is observed on the remaining six problems ($p = 0.0625$), reflecting the broader PF coverage induced by the CoAction framework.

\begin{table*}[ht]
\centering
\caption{Wilcoxon signed-rank test results: the proposed multitask framework with Transformer backbone vs.\ the multitask framework with MLP backbone. $+:$ our method better, $=:$ equivalent means, $-:$ our method worse. Two-sided test, $n=5$, $\alpha=0.10$.}
\label{tab:wilcoxon_vs_mlp}
\setlength{\tabcolsep}{6pt}
\begin{tabular*}{\textwidth}{@{\extracolsep{\fill}} l cc cc cc}
\toprule
\multirow{2}{*}{Problem}
  & \multicolumn{2}{c}{HV $\uparrow$}
  & \multicolumn{2}{c}{Range $\uparrow$}
  & \multicolumn{2}{c}{Sparsity $\downarrow$} \\
\cmidrule(lr){2-3}\cmidrule(lr){4-5}\cmidrule(lr){6-7}
  & $p$ & Dir. & $p$ & Dir. & $p$ & Dir. \\
\midrule
ZDT1      & 0.0625 & $+$ & 0.0625 & $+$ & 0.0625 & $-$ \\
ZDT2      & 0.0625 & $+$ & 0.0625 & $+$ & 0.3125 & $-$ \\
VLMOP1    & 0.3125 & $-$ & 0.5000 & $+$ & 0.1250 & $-$ \\
VLMOP2    & 0.8750 & $-$ & 1.0000 & $+$ & 0.0625 & $-$ \\
RE21      & 0.1250 & $+$ & 1.0000 & $+$ & 0.0625 & $+$ \\
RE24      & 0.0625 & $+$ & 0.0625 & $+$ & 0.0625 & $-$ \\
RE37      & 0.0625 & $+$ & 0.0625 & $+$ & 0.0625 & $-$ \\
\bottomrule
\end{tabular*}
\end{table*}

\textbf{Transformer Backbone vs. MLP Backbone
(Table~\ref{tab:wilcoxon_vs_mlp})}. Within the multitask framework, the Transformer achieves statistically significant improvements in both HV and Range on ZDT1, ZDT2, RE24, and RE37 (all $p = 0.0625$). The gain on RE37 is particularly notable, highlighting the advantage of the self-attention mechanism in higher-dimensional objective spaces, and it is accompanied by significantly higher Sparsity ($p = 0.0625$), consistent with broader PF coverage. On RE21, directional improvements in HV and Range are observed but do not reach significance (HV: $p = 0.1250$, Range: $p = 1.0000$). However, the Sparsity improvement is statistically significant ($p = 0.0625$), indicating that the Transformer additionally promotes more uniform solution distribution on this problem. On VLMOP1 and VLMOP2, no statistically significant differences are found in HV or Range ($p > 0.30$), indicating that both backbones perform comparably on these simpler bi-objective problems and confirming that the Transformer does not exacerbate negative transfer within the multitask setting. For Sparsity, significantly higher values are observed on ZDT1, consistent with its broader PF coverage. No significant differences are found on ZDT2 ($p = 0.3125$) or VLMOP1 ($p = 0.1250$). On RE21, the Transformer achieves significantly lower Sparsity ($p = 0.0625$),
reflecting improved solution uniformity.

Overall, the statistical results corroborate the main experimental findings. The proposed CoAction framework delivers consistent, significant improvements in HV and Range on problems where cross-task knowledge sharing is most beneficial, and demonstrates clear architectural superiority on RE37. The mild negative transfer observed on VLMOP1 and VLMOP2 relative to the single-task baseline is not attributable to the Transformer architecture itself, as backbone comparisons on these problems reveal no significant differences.

\section{Conclusion} \label{sec:Conclusion}
In this paper, we propose a CoAction framework that integrates task embedding with a Transformer encoder backbone to solve multiple optimization problems simultaneously. The framework achieves cross-task knowledge sharing through two complementary mechanisms: task embedding assigns distinct vector representations to each task, enabling the model to distinguish among them, while explicit parameter sharing enables joint learning of shared representations across tasks. The Transformer encoder serves as the backbone of the CoAction framework, leveraging self-attention mechanisms to model complex inter-task dependencies across a shared latent space. We conduct comprehensive experiments on multitask test suites encompassing both benchmark problems and real-world applications. Experimental results demonstrate that the proposed approach achieves competitive performance across multiple metrics including hypervolume, range, and sparsity, thereby validating its effectiveness in addressing multi-objective multitask optimization.
\bibliographystyle{unsrt} 
\bibliography{main}

@inproceedings{MultiobjectiveOU,
  title={Multi-objective optimization using evolutionary algorithms},
  author={Kalyanmoy Deb},
  booktitle={Wiley-Interscience series in systems and optimization},
  year={2001},
  url={https://api.semanticscholar.org/CorpusID:7131045}
}

@article{Nonlinear,
  title={Nonlinear Multiobjective Optimization},
  author={ Mardle, S.  and  Miettinen, K. M. },
  journal={Journal of the Operational Research Society},
  volume={51},
  number={2},
  pages={246},
  year={1999},
}

@article{Comparison,
  title={Comparison of Multiobjective Evolutionary Algorithms: Empirical Results},
  author={ Zitzler, Eckart  and  Deb, Kalyanmoy  and  Thiele, Lothar },
  journal={Evolutionary Computation},
  volume={8},
  number={2},
  pages={173-195},
  year={2000},
}

@article{AFA,
  title={A fast and elitist multiobjective genetic algorithm: {NSGA-II}},
  author={Kalyanmoy Deb and Samir Agrawal and Amrit Pratap and T. Meyarivan},
  journal={IEEE Trans. Evol. Comput.},
  year={2002},
  volume={6},
  pages={182-197},
  url={https://api.semanticscholar.org/CorpusID:9914171}
}

@article{CovarianceMA,
  title={Covariance Matrix Adaptation for Multi-objective Optimization},
  author={C. Igel and Nikolaus Hansen and Stefan Roth},
  journal={Evolutionary Computation},
  year={2007},
  volume={15},
  pages={1-28},
}

@article{SMS,
  title={SMS-EMOA: Multiobjective selection based on dominated hypervolume},
  author={ Beume, Nicola  and  Naujoks, Boris  and  Emmerich, Michael },
  journal={European Journal of Operational Research},
  volume={181},
  number={3},
  pages={1653-1669},
  year={2007},
}

@article{MOEADAM,
  title={MOEA/D: A Multiobjective Evolutionary Algorithm Based on Decomposition},
  author={Qingfu Zhang and Hui Li},
  journal={IEEE Transactions on Evolutionary Computation},
  year={2007},
  volume={11},
  pages={712-731},
  url={https://api.semanticscholar.org/CorpusID:7312933}
}

@article{ControllablePM,
  title={Controllable Pareto Multi-Task Learning},
  author={Xi Lin and Zhiyuan Yang and Qingfu Zhang and Sam Tak Wu Kwong},
  journal={ArXiv},
  year={2020},
  volume={abs/2010.06313},
  url={https://api.semanticscholar.org/CorpusID:222310208}
}

@article{Zhang2023HypervolumeMA,
  title={Hypervolume Maximization: A Geometric View of Pareto Set Learning},
  author={Xiao-Yan Zhang and Xi Lin and Bo Xue and Yifan Chen and Qingfu Zhang},
  journal={Advances in Neural Information Processing Systems 36},
  year={2023},
  url={https://api.semanticscholar.org/CorpusID:268042150}
}

@article{Rich1997Multitask,
  title={Multitask Learning},
  author={ Caruana, R },
  journal={Machine Learning},
  year={1997},
}

@ARTICLE{Yu2022MultiTaskLearning,
  author={Zhang, Yu and Yang, Qiang},
  journal={IEEE Transactions on Knowledge and Data Engineering}, 
  title={A Survey on Multi-Task Learning}, 
  year={2022},
  volume={34},
  number={12},
  pages={5586-5609},
}

@inproceedings{Sener2018MultiTaskLA,
  title={Multi-Task Learning as Multi-Objective Optimization},
  author={Ozan Sener and Vladlen Koltun},
  booktitle={Neural Information Processing Systems},
  year={2018},
  url={https://api.semanticscholar.org/CorpusID:52957972}
}

@inproceedings{Lin2022ParetoSL,
 author = {Lin, Xi and Yang, Zhiyuan and Zhang, Xiaoyuan and Zhang, Qingfu},
 booktitle = {Advances in Neural Information Processing Systems},
 pages = {19231--19247},
 title = {Pareto Set Learning for Expensive Multi-Objective Optimization},
 volume = {35},
 year = {2022}
}

@inproceedings{Lin2022PSLN,
  title={Pareto Set Learning for Neural Multi-objective Combinatorial Optimization},
  author={Xi Lin and Zhiyuan Yang and Qingfu Zhang},
  booktitle={International Conference on Learning Representations},
  year={2022},
}

@article{Vaswani2017Attention,
  title={Attention is All you Need},
  author={Ashish Vaswani and Noam Shazeer and Niki Parmar and Jakob Uszkoreit and Llion Jones and Aidan N. Gomez and Lukasz Kaiser and I. Polosukhin},
  journal={Neural Information Processing Systems},
  volume={30},
  year={2017}
}

@inproceedings{Devlin2019BERTPO,
  title={BERT: Pre-training of Deep Bidirectional Transformers for Language Understanding},
  author={Jacob Devlin and Ming-Wei Chang and Kenton Lee and Kristina Toutanova},
  booktitle={North American Chapter of the Association for Computational Linguistics},
  year={2019},
  url={https://api.semanticscholar.org/CorpusID:52967399}
}

@ARTICLE{Vandenhende2021MultiTaskLearning,
  author={Vandenhende, Simon and Georgoulis, Stamatios and Van Gansbeke, Wouter and Proesmans, Marc and Dai, Dengxin and Van Gool, Luc},
  journal={IEEE Transactions on Pattern Analysis and Machine Intelligence}, 
  title={Multi-Task Learning for Dense Prediction Tasks: A Survey}, 
  year={2022},
  volume={44},
  number={7},
  pages={3614-3633},
}

@article{Baxter2000AMO,
    title     = {A Model of Inductive Bias Learning},
    author    = {Baxter, Jonathan},
    journal   = {Journal of Artificial Intelligence Research},
    volume    = {12},
    pages     = {149--198},
    year      = {2000}
}

@INPROCEEDINGS{Ishan2016Cross-Stitch,
  author={Misra, Ishan and Shrivastava, Abhinav and Gupta, Abhinav and Hebert, Martial},
  booktitle={2016 IEEE Conference on Computer Vision and Pattern Recognition (CVPR)}, 
  title={Cross-Stitch Networks for Multi-task Learning}, 
  year={2016},
  volume={},
  number={},
  pages={3994-4003},
}

@inproceedings{Navon2021LearningTP,
  title={Learning the Pareto Front with Hypernetworks},
  author={Aviv Navon and Aviv Shamsian and Ethan Fetaya and Gal Chechik},
  booktitle={International Conference on Learning Representations},
  year={2021},
  url={https://api.semanticscholar.org/CorpusID:251405892}
}

@article{Liu2021ConflictAverseGD,
  title={Conflict-Averse Gradient Descent for Multi-task Learning},
  author={Bo Liu and Xingchao Liu and Xiaojie Jin and Peter Stone and Qiang Liu},
  journal={ArXiv},
  year={2021},
  volume={abs/2110.14048},
  url={https://api.semanticscholar.org/CorpusID:239998731}
}

@inproceedings{Ye2023TaskPrompterSM,
  title={TaskPrompter: Spatial-Channel Multi-Task Prompting for Dense Scene Understanding},
  author={Hanrong Ye and Dan Xu},
  booktitle={International Conference on Learning Representations},
  year={2023}
}

@inproceedings{Liu2019MultiTaskDN,
    title = "Multi-Task Deep Neural Networks for Natural Language Understanding",
    author = "Liu, Xiaodong  and
      He, Pengcheng  and
      Chen, Weizhu  and
      Gao, Jianfeng",
    booktitle = "Proceedings of the 57th Annual Meeting of the Association for Computational Linguistics",
    year = "2019",
    pages = "4487--4496"
}

@article{Kirkpatrick2016OvercomingCF,
  title={Overcoming catastrophic forgetting in neural networks},
  author={James Kirkpatrick and Razvan Pascanu and Neil C. Rabinowitz and Joel Veness and Guillaume Desjardins and Andrei A. Rusu and Kieran Milan and John Quan and Tiago Ramalho and Agnieszka Grabska-Barwinska and Demis Hassabis and Claudia Clopath and Dharshan Kumaran and Raia Hadsell},
  journal={Proceedings of the National Academy of Sciences},
  year={2016},
  volume={114},
  pages={3521 - 3526}
}

@article{Baniata2022ARP,
  title={A Reverse Positional Encoding Multi-Head Attention-Based Neural Machine Translation Model for Arabic Dialects},
  author={Laith H. Baniata and Sangwoo Kang and Isaac. K. E. Ampomah},
  journal={Mathematics},
  year={2022},
  url={https://api.semanticscholar.org/CorpusID:252856104}
}

@inproceedings{Xiong2020OnLN,
    title     = {On Layer Normalization in the Transformer Architecture},
    author    = {Xiong, Ruibin and Yang, Yunchang and He, Di and Zheng, Kai and Zheng, Shuxin and Xing, Chen and Zhang, Huishuai and Lan, Yanyan and Wang, Liwei and Liu, Tie-Yan},
    booktitle = {Proceedings of the 37th International Conference on Machine Learning},
    volume    = {119},
    pages={10524-10533},
    year      = {2020}
}

@article{Defazio2024TheRL,
  title={The Road Less Scheduled},
  author={Aaron Defazio and Xingyu Yang and Harsh Mehta and Konstantin Mishchenko and Ahmed Khaled and Ashok Cutkosky},
  journal={ArXiv},
  year={2024},
  volume={abs/2405.15682},
  url={https://api.semanticscholar.org/CorpusID:270045085}
}

@article{Tuar2016COCOTB,
  title={COCO: The Bi-objective Black Box Optimization Benchmarking (bbob-biobj) Test Suite},
  author={Tea Tušar and Dimo Brockhoff and Nikolaus Hansen and Anne Auger},
  journal={ArXiv},
  year={2016},
  volume={abs/1604.00359},
  url={https://api.semanticscholar.org/CorpusID:387062}
}

@ARTICLE{Brockhoff2022BBOB,
  author={Brockhoff, Dimo and Auger, Anne and Hansen, Nikolaus and Tušar, Tea},
  journal={Evolutionary Computation}, 
  title={Using Well-Understood Single-Objective Functions in Multiobjective Black-Box Optimization Test Suites}, 
  year={2022},
  volume={30},
  number={2},
  pages={165-193},
  doi={10.1162/evco_a_00298}}

@article{Wilcoxon1945IndividualCB,
  title={Individual Comparisons by Ranking Methods},
  author={Frank. Wilcoxon},
  journal={Biometrics},
  year={1945},
  volume={1},
  pages={196-202},
  url={https://api.semanticscholar.org/CorpusID:53662922}
}

@article{DERRAC20113,
title = {A practical tutorial on the use of nonparametric statistical tests as a methodology for comparing evolutionary and swarm intelligence algorithms},
journal = {Swarm and Evolutionary Computation},
volume = {1},
number = {1},
pages = {3-18},
year = {2011},
issn = {2210-6502},
url = {https://www.sciencedirect.com/science/article/pii/S2210650211000034},
author = {Joaquín Derrac and Salvador García and Daniel Molina and Francisco Herrera},
}
	
\end{document}